%% file: main.tex
\documentclass{article} % For LaTeX2e

\usepackage[table]{xcolor}

\usepackage{colm2024_conference}

\usepackage{microtype}
\usepackage{hyperref}
\usepackage{url}
\usepackage{booktabs}
\definecolor{darkblue}{rgb}{0, 0, 0.5}
\hypersetup{colorlinks=true, citecolor=darkblue, linkcolor=darkblue, urlcolor=darkblue}
\usepackage{graphicx}
\usepackage{fancyhdr}
\input{math_commands.tex}

\usepackage{diagbox}
\usepackage{tabularx}
\usepackage{subcaption}
\usepackage{multirow} 

\usepackage{pifont}
\usepackage{listings}
\usepackage[most]{tcolorbox}
\newcommand{\eg}{\textit{e.g.}}

\newcommand{\tablestyle}[2]{\setlength{\tabcolsep}{#1}\renewcommand{\arraystretch}{#2}\centering\footnotesize}

\definecolor{lightgray}{gray}{0.95}
\lstdefinestyle{prompt}{
    basicstyle=\ttfamily\fontsize{7pt}{8pt}\selectfont,
    frame=none,
    breaklines=true,
    backgroundcolor=\color{lightgray},
    breakatwhitespace=true,
    breakindent=0pt,
    escapeinside={(*@}{@*)},
    numbers=none,
    numbersep=5pt,
    xleftmargin=5pt,
}
\lstdefinestyle{prompt_json}{
  basicstyle=\ttfamily,
  keywordstyle=\color{blue},
  stringstyle=\color{orange},
  commentstyle=\color{green},
  frame=single,
  rulecolor=\color{black},
  breakatwhitespace=false,
  breaklines=true,
  captionpos=b,
  keepspaces=true,
  showspaces=false,
  showstringspaces=false,
  showtabs=false,
  tabsize=2
}
\lstdefinestyle{prompt_dialog}{
  language=[Objective]C,
  basicstyle=\ttfamily,
  keywordstyle=\color{blue},
  commentstyle=\color{green},
  stringstyle=\color{orange},
  frame=single,
  rulecolor=\color{black},
  breakatwhitespace=false,
  breaklines=true,
  captionpos=b,
  keepspaces=true,
  showspaces=false,
  showstringspaces=false,
  showtabs=false,
  tabsize=2
}

\tcbset{
  aibox/.style={
    top=10pt,
    colback=white,
    colframe=black,
    colbacktitle=black,
    enhanced,
    center,
    attach boxed title to top left={yshift=-0.1in,xshift=0.15in},
    boxed title style={boxrule=0pt,colframe=white,},
  }
}
\newtcolorbox{AIbox}[2][]{aibox, title=#2,#1}

\usepackage[capitalize]{cleveref}

\crefname{section}{Sec.}{Secs.}
\Crefname{section}{Section}{Sections}
\Crefname{table}{Table}{Tables}
\crefname{table}{Tab.}{Tabs.}
\crefname{figure}{Fig.}{Figs.}
\Crefname{figure}{Figure}{Figures}
\crefname{appendix}{Appx.}{Appxs.}
\Crefname{appendix}{Appendix}{Appendixs}
\crefname{subsection}{Sec.}{Secs.}

\title{InternLM2 Technical Report}

\author{
\\
\parbox{\linewidth}{
\textbf{Zheng Cai, Maosong Cao, Haojiong Chen, Kai Chen, Keyu Chen, Xin Chen, Xun Chen, Zehui Chen, Zhi Chen, Pei Chu, Xiaoyi Dong, Haodong Duan, Qi Fan, Zhaoye Fei, Yang Gao, Jiaye Ge, Chenya Gu, Yuzhe Gu, Tao Gui, Aijia Guo, Qipeng Guo, Conghui He, Yingfan Hu, Ting Huang, Tao Jiang, Penglong Jiao, Zhenjiang Jin, Zhikai Lei, Jiaxing Li, Jingwen Li, Linyang Li, Shuaibin Li, Wei Li, Yining Li, Hongwei Liu, Jiangning Liu, Jiawei Hong, Kaiwen Liu, Kuikun Liu, Xiaoran Liu, Chengqi Lv, Haijun Lv, Kai Lv, Li Ma, Runyuan Ma, Zerun Ma, Wenchang Ning, Linke Ouyang, Jiantao Qiu, Yuan Qu, Fukai Shang, Yunfan Shao, Demin Song, Zifan Song, Zhihao Sui, Peng Sun, Yu Sun, Huanze Tang, Bin Wang, Guoteng Wang, Jiaqi Wang, Jiayu Wang, Rui Wang, Yudong Wang, Ziyi Wang, Xingjian Wei, Qizhen Weng, Fan Wu, Yingtong Xiong, Chao Xu, Ruiliang Xu, Hang Yan, Yirong Yan, Xiaogui Yang, Haochen Ye, Huaiyuan Ying, Jia Yu, Jing Yu, Yuhang Zang, Chuyu Zhang, Li Zhang, Pan Zhang, Peng Zhang, Ruijie Zhang, Shuo Zhang, Songyang Zhang, Wenjian Zhang, Wenwei Zhang, Xingcheng Zhang, Xinyue Zhang, Hui Zhao, Qian Zhao, Xiaomeng Zhao, Fengzhe Zhou, Zaida Zhou, Jingming Zhuo, Yicheng Zou, Xipeng Qiu, Yu Qiao, Dahua Lin} }\\   
\\
\textbf{Shanghai AI Laboratory} \\
\textbf{SenseTime Group} \\
\textbf{The Chinese University of Hong Kong}\\
\textbf{Fudan University} \\ 
\texttt{internlm@pjlab.org.cn}
}

\colmfinalcopy 

\begin{document}

\maketitle

\begin{abstract}
The evolution of Large Language Models (LLMs) like ChatGPT and GPT-4 has sparked discussions on the advent of Artificial General Intelligence (AGI). However, replicating such advancements in open-source models has been challenging. This paper introduces InternLM2, an open-source LLM that outperforms its predecessors in comprehensive evaluations across 6 dimensions and 30 benchmarks, long-context modeling, and open-ended subjective evaluations through innovative pre-training and optimization techniques. The pre-training process of InternLM2 is meticulously detailed, highlighting the preparation of diverse data types including text, code, and long-context data. InternLM2 efficiently captures long-term dependencies, initially trained on 4k tokens before advancing to 32k tokens in pre-training and fine-tuning stages, exhibiting remarkable performance on the 200k ``Needle-in-a-Haystack" test. InternLM2 is further aligned using Supervised Fine-Tuning (SFT) and a novel Conditional Online Reinforcement Learning from Human Feedback (COOL RLHF) strategy that addresses conflicting human preferences and reward hacking. By releasing InternLM2 models in different training stages and model sizes, we provide the community with insights into the model's evolution.
\end{abstract}

\newpage

\tableofcontents 

\section{Introduction}
Since the introduction of ChatGPT and GPT-4~\citep{DBLP:journals/corr/abs-2303-08774}, Large Language Models (LLMs) have surged in popularity across the academic and industrial spheres. Models trained on billions of tokens have demonstrated profound empathy and problem-solving capabilities, leading to widespread speculation that the era of Artificial General Intelligence (AGI) may soon be upon us. Despite this enthusiasm, the path to developing models with capabilities comparable to those of ChatGPT or GPT-4 remains elusive. The open-source community has been working diligently to bridge the gap between proprietary LLMs and their open-source counterparts. In the past year, several notable open-source LLMs, such as LLaMA~\citep{DBLP:journals/corr/abs-2302-13971,DBLP:journals/corr/abs-2307-09288}, Qwen~\citep{DBLP:journals/corr/abs-2309-16609}, Mistral~\citep{DBLP:journals/corr/abs-2310-06825}, and Deepseek~\citep{DBLP:journals/corr/abs-2401-02954}, have made significant strides. In this paper, we introduce InternLM2, a new Large Language Model that outperforms the previously mentioned models.

The development of Large Language Models (LLMs) encompasses several main phases: pre-training, Supervised Fine-Tuning (SFT), and Reinforcement Learning from Human Feedback (RLHF)~\citep{DBLP:conf/nips/Ouyang0JAWMZASR22}. Pre-training is chiefly based on leveraging a vast corpus of natural text, amassing trillions of tokens. This phase is aimed at equipping LLMs with a broad repository of knowledge and fundamental skills. The quality of data is considered the most crucial factor during pre-training. However, technical reports on LLMs~\citep{DBLP:journals/corr/abs-2302-13971,DBLP:journals/corr/abs-2307-09288,DBLP:journals/corr/abs-2309-16609,DBLP:journals/corr/abs-2401-02954} in the past have seldom addressed the processing of pre-training data. InternLM2 extensively details how it prepares text, code, and long-context data for pre-training.

How to effectively extend the context length of LLMs is currently a hot research topic, since many downstream applications, such as Retrieval-Augmented Generation (RAG)~\citep{DBLP:journals/corr/abs-2312-10997} and agents~\citep{DBLP:journals/corr/abs-2309-07864}, rely on long contexts. 
InternLM2 first employs Group Query Attention (GQA) to enable a smaller memory footprint when inferring long sequences. In the pre-training phase, we initially train InternLM2 with 4k context texts, then transit the training corpus to high-quality 32k texts for further training. Upon completion, through positional encoding extrapolation~\citep{dynamicNTK}, InternLM2 achieves commendable performance in the ``Needle-in-a-Haystack'' test within 200k contexts.

Following long-context pre-training, we utilize supervised fine-tuning (SFT) and reinforcement learning from human feedback (RLHF) to ensure the model adheres well to human instructions and aligns with human values. Notably, we also construct corresponding 32k data during these processes to further improve the long-context processing capability of InternLM2. Besides, we introduce \textbf{CO}nditional \textbf{O}n\textbf{L}ine RLHF (COOL RLHF), which adopts a conditional reward model to reconcile diverse but potentially conflicting preferences and executes Proximal Policy Optimization (PPO) over multiple rounds to mitigate emerging reward hacking in each phase. To elucidate the impact of RLHF within the community, we also release models in their pre- and post-RLHF stages, named InternLM2-Chat-\{size\}-SFT and InternLM2-Chat-\{size\}, respectively.

Our contributions are twofold, highlighting not only the model's exceptional performance across diverse benchmarks but also our comprehensive approach to its development in different stages. Key points include 

\begin{enumerate}
  \item \textbf{Open-Sourcing InternLM2 with Exceptional Performance:} We have open-sourced models of various sizes, including 1.8B, 7B, and 20B, all of which perform well in both subjective and objective evaluations. Additionally, we have released models from different stages to facilitate community analysis of changes post-SFT and RLHF training.
  \item \textbf{Designed with a 200k Context Window:} InternLM2 exhibits impressive long-context performance, nearly perfectly identifying all ``needles" in the ``Needle-in-a-Haystack" experiment with a 200k context. Furthermore, we provide experience of training long-context LLMs across all stages, including pretraining, SFT, and RLHF.
  \item \textbf{Comprehensive Data Preparation Guidance:} We have elaborated on the preparation of data for LLMs, including pre-training data, domain-specific enhancement data, SFT data, and RLHF data. These details will benefit the community in better training of LLMs.
  \item \textbf{Innovative RLHF Training Techniques:} We introduced Conditional Online RLHF (COOL RLHF) to harmonize various preferences, significantly enhancing the performance of InternLM2 in various subjective dialogue evaluations. We have also conducted a preliminary analysis and comparison of RLHF's subjective and objective results to offer the community insights into RLHF.
\end{enumerate}

\section{Infrastructure}
In this part, we introduce the training framework InternEvo which was used during pretraining, SFT and RLHF.

\subsection{InternEvo}

We utilize InternEvo, an efficient and lightweight pretraining framework, for model training.  This framework enables us to scale model training across thousands of GPUs. This is achieved through a combination of data, tensor~\citep{tensorparallelism2019}, sequence~\citep{sequenceparallelism2023}, and pipeline~\citep{pipelineparallelism2019} parallelism. To further enhance GPU memory efficiency, InternEvo incorporates various Zero Redundancy Optimizer (ZeRO)~\citep{rajbhandari2020zero} strategies, significantly reducing the memory footprint required for training. In addition, to enhance hardware utilization, we incorporate the FlashAttention technique~\citep{DBLP:journals/corr/abs-2307-08691} and mixed-precision training~\citep{narang2017mixed} with BF16.

\begin{figure}[h]
    \centering
    \includegraphics[width=0.9\textwidth]{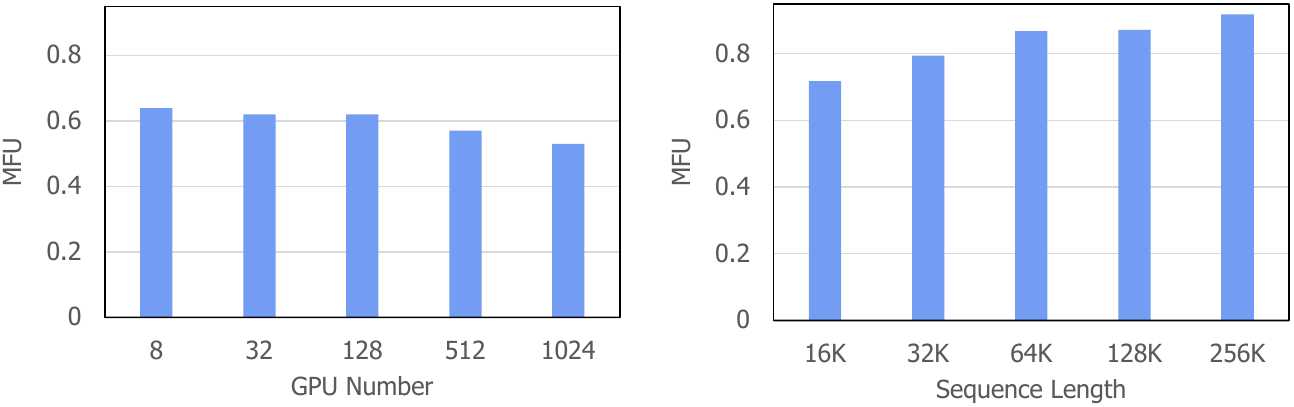}
    \caption{Model FLOPs Utilization (MFU) of training InternLM-7B with InternEvo. We benchmark training performance using a sequence length of 4096 tokens with varying GPU numbers, and benchmark training performance on 128 GPUs with varying sequence lengths.}
    \label{fig:mfu}
\end{figure}

InternEvo demonstrates strong scaling performance when training InternLM across thousands of GPUs. As shown in Figure \ref{fig:mfu}, when training InternLM-7B  on 8 GPUs with a global batch size of 4 million tokens, InternEvo achieves 64\% Model FLOPs Utilization (MFU). Scaling up the training to 1024 GPUs, InternEvo maintains an impressive 53\% MFU with the same global batch size. This level of scaling performance is particularly challenging to attain, as the batch size remains constant and the computation-to-communication ratio decreases with an increased number of GPUs. In contrast, DeepSpeed~\citep{rasley2020deepspeed} only manages to achieve around 36\% MFU when training the InternLM-7B on 1024 GPUs using ZeRO-1~\citep{rajbhandari2020zero} and MiCS~\citep{zhang2022mics}. 

InternEvo also exhibits strong scaling of sequence length, supporting  $256,000$ tokens when training LLMs of varying sizes. For instance, when training the InternLM-7B model with a sequence length of $256,000$ tokens, InternEvo can achieve nearly 88\% MFU. In comparison, DeepSpeed-Ulysses and Megatron-LM only reach about 65\% MFU. This improvement in training performance is also observed with LLMs of larger sizes, such as models with 30 billion or 70 billion parameters.

\paragraph{Reducing Communication Overhead}
A trade-off exists between memory utilization and communication cost in distributed LLM training. Initially, the communication cost can be effectively reduced by diminishing the communication scale. This involves limiting communications to a smaller group of GPUs, potentially within the same node, which mitigates the overall communication cost. Building upon this principle,  InternEvo addresses communication challenges by implementing a suite of adaptive sharding techniques to achieve  strong scaling performance~\citep{chen2024amsp}. These include Full-Replica, Full-Sharding, and Partial-Sharding, which allow each component of the model states—parameters, gradients, and optimizer states—to independently select the most appropriate sharding approach and device mesh configuration. This flexibility facilitates a more nuanced distribution of model states across the GPU infrastructure. InternEvo also introduces an optimization framework designed to identify the most efficient sharding factors. This aims to minimize communication expenses while adhering to the memory constraints of the GPU.

\paragraph{Communication-Computation Overlap}
Further reducing communication overhead, InternEvo strategically coordinates communication and computation to optimize overall system performance. When employing parameter sharding, the model's entire parameters are distributed across multiple GPUs to conserve GPU memory. During each forward and backward pass of every micro-batch in a training step, InternEvo efficiently pre-fetches the complete parameter set for upcoming layers through AllGather, while concurrently computing the current layer. The generated gradients undergo synchronization within the parameter sharding group through ReduceScatter and are subsequently synchronized across parameter sharding groups using AllReduce. These communication processes are skillfully overlapped with the backward computation, maximizing efficiency in the training pipeline. In the case of optimizer state sharding, where the GPU broadcasts updated parameters within the sharding group through Broadcast, InternEvo employs a strategic overlap with the forward computation of the next training step. These innovative overlap approaches effectively balance communication overhead and computation execution time, resulting in a significant enhancement in overall system performance.

\paragraph{Long-Sequence Training}
One of the primary challenges in long-sequence training is the trade-off between computation speed and communication overhead. InternEvo breaks down GPU memory management into a hierarchical space with four parallel dimensions—data, tensor, sequence, and pipeline—and three sharding dimensions—parameter, gradient, and optimizer state~\citep{InternEvo2024}. We conduct a thorough analysis of memory and communication costs for each dimension, utilizing an execution simulator to identify and implement the optimal parallelization strategy. The optimal execution plan can be automatically searched based on the training scale, sequence length, model size, and  batch size. With this execution plan, InternEvo exhibits the capability to handle long contexts (up to 1 million tokens) during training. InternEvo also implements memory management techniques to reduce GPU memory fragmentation, a common issue in long-sequence training scenarios. It uses a memory pool for unified memory management and introduces a defragmentation technique that proactively consolidates small memory chunks to prevent out-of-memory errors.

\paragraph{Fault Tolerance}

We have also addressed the challenges of efficiently training LLMs in GPU datacenters, which often face issues such as frequent hardware failures, complex parallelization strategies, and imbalanced resource utilization. To tackle these issues, we conducted an in-depth characterization study of a six-month LLM development workload trace from our GPU datacenter~\citep{hu2024characterization}. This study identified discrepancies between LLMs and previous deep learning workloads and explored resource utilization patterns and job failure impacts. Based on our analysis, we introduced two system efforts: a fault-tolerant pretraining system that enhances fault tolerance through LLM-involved failure diagnosis and automatic recovery, and a decoupled scheduling system for evaluation tasks that provides timely model performance feedback.
In our implementation, we have incorporated an asynchronous saving mechanism that regularly archives model weights and optimizer states to distributed file and object storage at predefined intervals. Throughout the training process, each GPU first saves its model states in local storage and subsequently asynchronously uploads this data to the remote distributed storage system. This dual-step process ensures that, in the event of occasional hardware or network failures automatically detected by our system, only a minimal amount of training progress is lost. To optimize storage space, we systematically transfer these temporary model checkpoints from hot storage to cost-effective cold storage. Furthermore, our system is designed to seamlessly resume model training even when the parallelization configuration is altered, providing flexibility and continuity in the training pipeline.

\paragraph{Interactive Training}

The efficiency of InternEvo has also been successfully demonstrated in the Reinforcement Learning from Human Feedback (RLHF) stage, where multiple LLMs are deployed for interactive training. For instance, in the Proximal Policy Optimization (PPO) process, we utilize four equally-sized models and train two of them; InternEvo enables each model to be executed at its optimal configuration. To enhance the coordination of multiple models, we have developed an innovative RLHF framework that builds upon InternEvo and Ray. This framework is characterized by its flexibility and scalability, allowing it to perform effectively at scale. It is capable of integrating with various LLM execution engines and supporting diverse algorithmic designs. For a comprehensive description of the "Alignment" concept, please refer to Section \ref{sec:Alignment}.

\subsection{Model Structure}
\begin{figure}[!t]
  \centering
  \includegraphics[width=\textwidth]{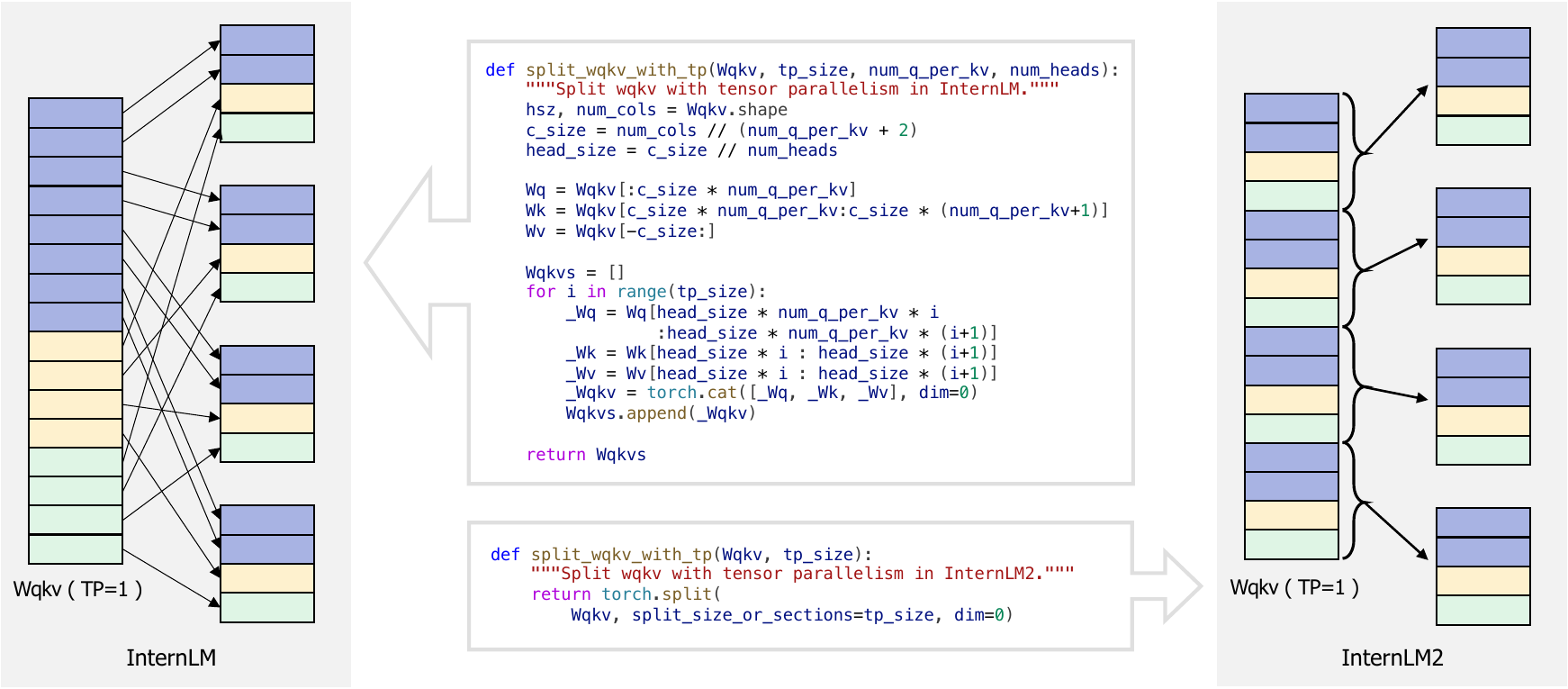}
  \caption{Different weight matrix layouts cause different complexity when changing the tensor parallelism~(TP) size.} \label{fig:Wqkv_tp}
\end{figure}

Transformer~\citep{DBLP:conf/nips/VaswaniSPUJGKP17} has been predominantly used as the backbone for past Large Language Models (LLMs) due to its excellent parallelization capabilities, which fully leverage the power of GPUs~\citep{DBLP:conf/nips/BrownMRSKDNSSAA20,DBLP:journals/jmlr/ChowdheryNDBMRBCSGSSTMRBTSPRDHPBAI23,DBLP:conf/iclr/ZengLDWL0YXZXTM23}. LLaMA~\citep{DBLP:journals/corr/abs-2302-13971} builds on  Transformer architecture by replacing LayerNorm~\citep{DBLP:journals/corr/BaKH16} with RMSNorm~\citep{DBLP:conf/nips/ZhangS19a} and setting the activation function to SwiGLU~\citep{DBLP:journals/corr/abs-2002-05202}, resulting in improved training efficiency and performance.  
Since the unveiling of LLaMA~\citep{DBLP:journals/corr/abs-2302-13971}, the community has vigorously engaged in augmenting the ecosystem built around the LLaMA architecture. This includes advancements in high-efficiency inference~\citep{llama.cpp} and operator optimization~\citep{DBLP:journals/corr/abs-2307-08691}, among others. To ensure our model, InternLM2, aligns seamlessly with this well-established ecosystem, alongside other renowned LLMs, such as Falcon~\citep{DBLP:journals/corr/abs-2311-16867}, Qwen~\citep{DBLP:journals/corr/abs-2309-16609}, Baichuan~\citep{DBLP:journals/corr/abs-2309-10305}, Mistral~\citep{DBLP:journals/corr/abs-2310-06825}, we opt to adhere to the structural design principles of LLaMA. In pursuit of enhancing efficiency, we consolidate the $W_k$, $W_q$, and $W_v$ matrices, which has resulted in a training acceleration of over 5\% during the pre-training phase. Moreover, to better support diverse tensor parallelism (tp) transformations, we have reconfigured the matrix layout. Rather than stacking the $W_k$, $W_q$, and $W_v$ matrices in a straightforward manner, we adopt an interleaving approach for each head's $W_k$, $W_q$, and $W_v$, as depicted in Figure~\ref{fig:Wqkv_tp}. This design modification facilitates the adjustment of the tensor parallel size through either splitting or concatenating the matrices along their last dimension, thereby enhancing the model's flexibility in varying distributed computing environments. InternLM2 aims to infer beyond 32K context, therefore, InternLM2 series models all choose Grouped-Query Attention~(GQA)~\citep{DBLP:conf/emnlp/AinslieLJZLS23}, so that it can infer both in high speed and low GPU memory with very long contexts.  

\section{Pre-train}
We introduce pre-training data, pre-training settings, and three pre-training phases in this part. 

\subsection{Pre-training data}
The pre-training of LLMs is critically shaped by data, which presents a multifaceted challenge. It encompasses handling sensitive data, covering comprehensive knowledge, and balancing efficiency and quality. In this section, we will depict our data processing pipeline for preparing general domain textual data, programming language-related data, and long textual data.

\subsubsection{Text Data}
The text data in our pre-training dataset can be categorized by source into web pages, papers, patents, and books. To transform these sources into a pre-training dataset, we first standardize all data into a specified format, categorize them by type and language, and store them in JSON Lines (jsonl) format. Then, for all data, we apply several processing steps including rule-based filtering, data deduplication, safety filtering, and quality filtering. This results in a rich, safe, and high-quality text dataset.

\paragraph{Data Source Distribution}

We have statistically analyzed the number of documents, data storage size, and data Bytes proportion in the pre-training dataset according to their sources, as shown in Table \ref{tab:data-summary}. Among them, the Chinese and English data from web pages account for 86.46\% of the total, making it the primary source. Although the data volume from other sources is relatively small, such as books and technical literature(abbreviated as techlit), the average document length is longer and the content quality is relatively higher, making them equally important.

\begin{table}[h]
    \centering
    \begin{tabular}{lrrr}
    \toprule
    Source      & {Docs (M rows)} & {Bytes (GB)} &  {Bytes-percent} \\
    \midrule
    en-books    &    0.50       &  220.14   &   1.63\%       \\
    en-techlit  &   59.27       &  576.48   &   4.27\%       \\
    en-webpages & 3614.07       & 9129.39   &  67.51\%       \\
    zh-books    &    0.71       &  366.82   &   2.71\%       \\ 
    zh-techlit  &   89.59       &  668.19   &   4.94\%       \\
    zh-webpages &  928.94       & 2562.86   &  18.95\%       \\ 
    \bottomrule
    \end{tabular}
    \caption{Summary of the pre-train data from different sources}\label{tab:data-summary}
\end{table}

\begin{figure}[h]
    \centering
    \includegraphics[width=0.9\textwidth]{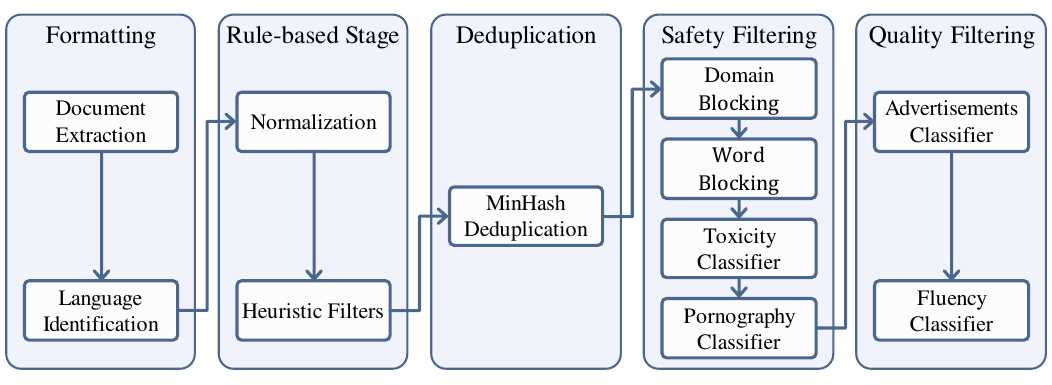}
    \caption{Data Process Pipeline}
    \label{fig:text-pipline}
\end{figure}

\paragraph{Data Process Pipeline}
The data processing pipeline used in this work is shown in Figure \ref{fig:text-pipline}. The entire data processing pipeline first standardizes data from different sources to obtain \textbf{Format data}. Then, heuristic statistical rules are used for data filtering to get \textbf{Clean data}. Next, the Locality-Sensitive Hashing (LSH) method is used for data deduplication to obtain \textbf{Dedup data}. We then apply a composite safety strategy to filter the data, resulting in \textbf{Safe data}. We have adopted different quality filtering strategies for data from various sources, ultimately obtaining \textbf{High-quality pre-training data}.

\paragraph{Data Formatting}
We will detail the data processing pipeline using web page data as an example. Our web page data mainly comes from Common Crawl\footnote{https://commoncrawl.org/}. Firstly, we need to decompress the original Warc format files and use Trafilatura~\citep{DBLP:conf/acl/Barbaresi21} for HTML parsing and main text extraction. Then, we use the pycld2\footnote{https://pypi.org/project/pycld2/} library for language detection and classification of the main text. Finally, we assign a unique identifier to the data and store it in jsonl (JSON lines) format and obtained \textbf{Format data}.

\paragraph{Rule-based Stage}
Web page data randomly extracted from the internet often contains a large amount of low-quality data, such as parsing errors, formatting errors, and non-natural language text. A common practice is to design rule-based regularization and filtering methods to modify and filter the data, as seen in Gopher~\citep{DBLP:journals/corr/abs-2112-11446}, C4~\citep{DBLP:conf/emnlp/DodgeSMAIGM021}, and RefinedWeb~\citep{DBLP:journals/corr/abs-2306-01116}. Based on our observations of the data, we have designed a series of heuristic filtering rules that focus on anomalies in separation and line breaks, frequency of abnormal characters, and distribution of punctuation marks. By applying these filters, we obtained \textbf{Clean data}.

\paragraph{Deduplication}
A large amount of duplicate texts exist on the Internet, which can negatively impact model training. Therefore, we employed a method based on Locality-Sensitive Hashing (LSH) to perform fuzzy deduplication on the data. More specifically, we used the MinHash method~\citep{DBLP:conf/sequences/Broder97}, establishing signatures with 128 hash functions on the 5-gram of the documents, and using 0.7 as the threshold for deduplication. We aimed to retain the most recent data, that is, prioritizing data with larger CC dumps numbers. We obtained the \textbf{Dedup data} after LSH deduplication.

\paragraph{Safety Filtering}
The internet is rife with toxic and pornographic content, the use of which for model training can negatively impact performance and increase the likelihood of unsafe content generation. Therefore, we employed a comprehensive safety strategy combining ``domain blocking'', ``word blocking'', ``pornography classifier'', and ``toxicity classifier'' to filter the data. Specifically, we constructed a block domain list comprising approximately 13M unsafe domains and a block word list containing 36,289 unsafe words for preliminary data filtering. Given that word blocking might inadvertently exclude a large amount of data, we opted for a cautious approach in compiling the block word list.

To further improve the detection rate of unsafe content, we fine-tuned the BERT model using the ``Toxic Comment Classification Challenge" dataset from Kaggle, resulting in a toxicity classifier. We sampled some data from \textbf{Dedup data} and annotated it using the Perspective API\footnote{https://perspectiveapi.com/} to create a pornography classification dataset. We then fine-tuned the BERT model with this dataset, yielding a pornography classifier. Finally, we used these two classifiers for secondary filtering of the data, filtering out data with scores below the threshold, resulting in \textbf{Safe data}.

\paragraph{Quality Filtering}
Compared to sources such as books, papers, and patents, internet-sourced data contains a significant amount of low-quality content. Based on our observations, the primary causes for this low-quality content are twofold: 1. The internet is rife with marketing advertisements, which tend to be repetitive and carry little information. 2. Many web pages consist of lists of article abstracts or product descriptions, resulting in extracted text that is difficult to read and lacks logical coherence.

To filter out these low-quality contents, we first organized manual data annotation. For the advertisements classification task, annotators were asked to identify whether a piece of data contains advertising content (both overall and partial advertisements are marked as low quality). For the fluency classification task, annotators were asked to rate the data on four dimensions: consistency, noise, information content, and grammar, resulting in a comprehensive fluency score. We then fine-tuned the BERT model using the manually annotated data, obtaining an advertisements classifier and a fluency classifier. Finally, we used these two classifiers for secondary filtering of the data, filtering out data with scores below the threshold, resulting in \textbf{High-quality pre-train data}.

\subsubsection{Code Data}
Programming is a crucial skill for a LLM, offering support for a variety of downstream applications, such as coding assistance, software development, and building tool-use agents. Moreover, \citet{DBLP:journals/corr/abs-2402-00838} indicate the possibility of enhancing reasoning capabilities by training on code data, as code is generally well-structured, rigorous, and predictable than natural language.

\paragraph{Data Source Distribution}

\begin{figure}[h]
    \centering
    \includegraphics[width=0.95\linewidth]{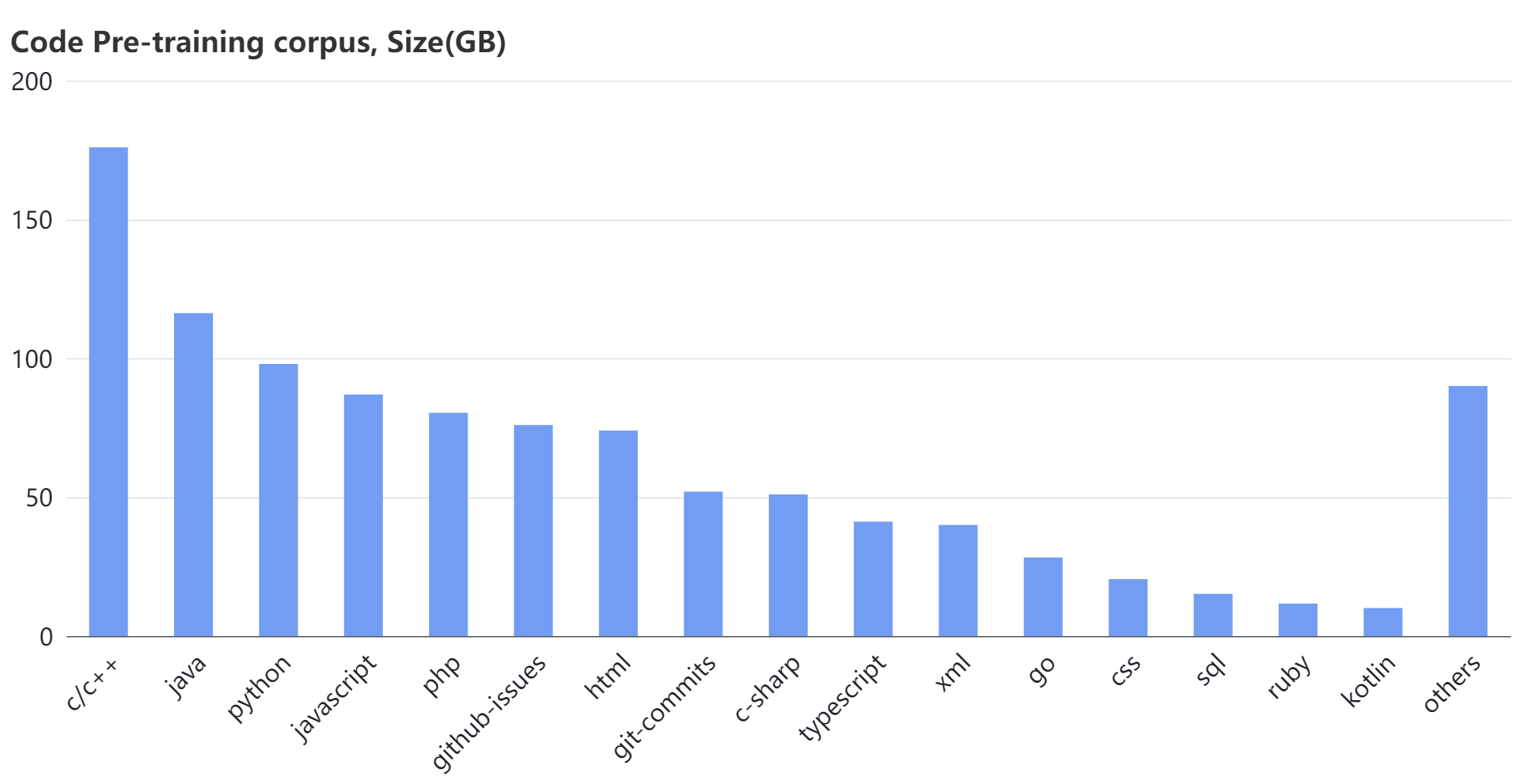}
    \caption{Statistics of code data in our pre-training corpus.}
    \label{fig:code_data_source}
\end{figure}

We collect data from various sources, including direct crawling from GitHub, public datasets, and online resources related to coding and programming, like Q\&A forums, tutorial sites, and API documentation, the statistics is shown in Figure~\ref{fig:code_data_source}.

Table~\ref{tab:code_data_level} reflects the data quality assessment based on a scoring model we trained. High-quality data will have a higher sampling weight and can undergo multiple training iterations in the pre-training phase. Moderate-quality data has a normal sampling weight and is typically trained once. Low-quality data are excluded, as our empirical findings affirm that removing them is vital for optimizing model performance and ensuring training stability despite their proportion being relatively small.

\begin{table}[h]
    \centering
    \begin{tabular}{cc|cc|cc}
    \toprule
    \multicolumn{2}{c}{High} & \multicolumn{2}{c}{Moderate} & \multicolumn{2}{c}{Low} \\
     Bytes (GB) & Percentage & Bytes (GB) & Percentage & Bytes (GB) & Percentage\\
    \midrule
    105.6 & 16.8\% & 440.1 & 69.9\% & 83.85 & 13.3\% \\
\bottomrule
    \end{tabular}
    \caption{Statistics of code data quality based on a learnable classifier, where high-quality data will be trained multiple times, moderate-quality data will be trained once, and low-quality will be dropped. The data of low-resource programming languages will be preserved and not taken account in the statistics.}
    \label{tab:code_data_level}
\end{table}

\paragraph{Format Cleaning}

All data is converted to a unified markdown format. Nevertheless, a very small fraction of the data still exhibited corrupted HTML or XML formats. We applied a set of heuristic rules to minimize these occurrences, though we did not invest too much effort in format purification. Markdown was selected for its simplicity—minimizing the token overhead for formatting—and its compatibility with interleaving code and natural language. The real format used for the pre-training is more complex, involving the concatenation of multiple code files based on their dependencies. The main idea is to utilize the interleaving data, which is pivotal for teaching the model about programming. This point is also mentioned in recent studies~\citep{DBLP:journals/corr/abs-2401-14196}.

\paragraph{Data Deduplication}

Deduplicating code data is similar to processing natural language except for tokenization, which impacts hyperparameter selection. For instance, Python examples use two spaces, four spaces, or a tab character to signify indentation. A conventional whitespace tokenizer, or one tailored for natural language, might mistakenly assess these samples as different data, which is inaccurate. Our insight is that an effective tokenizer is essential for applying a universal deduplication strategy. Although recent studies have explored fine-grained deduplication at the paragraph or line level, our approach remains at the file level to preserve context integrity.

\paragraph{Quality Filtering}

Quality is a pivotal yet nebulous aspect of pre-training in LLM research, primarily due to the difficulty in quantifying its impact on model performance regarding the scale. We adopted a hybrid, multi-stage filtering process including rule- and model-based scorers. Rule-based scorers are heuristic and varied, though we discovered that code style is not a reliable quality metric and can misclassify too many codes as low-quality. For the model-based scoring, we evaluated several backbone models, training them with roughly 50,000 labeled samples. However, we observed that the correlation between scorer assessments and human judgments varied across languages, and enlarging the training set did not substantially enhance scorer accuracy. Consequently, we only employed model-based scoring for languages where the model predictions align well with human evaluations on a human-annotated validated set.

\begin{figure}[t]
    \centering
    \includegraphics[width=0.7\linewidth]{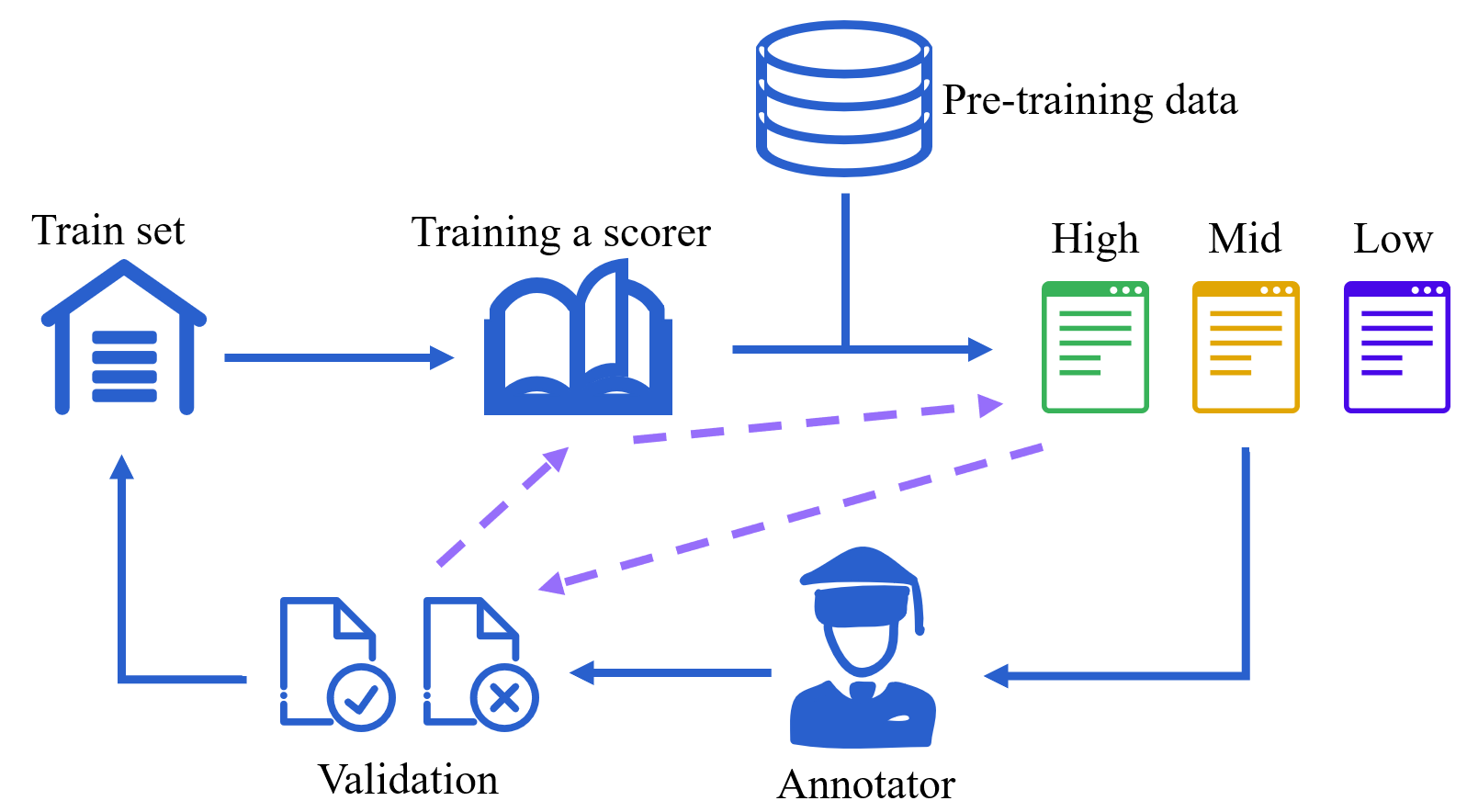}
    \caption{The pipeline of iterative refinement annotation process for a code quality classifier.}
    \label{fig:code_anno_pipeline}
\end{figure}
In order to obtain reliable annotations of our model-based scorer, we introduce an iterative annotation process (illustrated in Figure~\ref{fig:code_anno_pipeline}) to address the challenge that the definition of code quality data is vague. Identifying code that would be helpful for teaching an LLM is also non-trivial for human experts, for instance, a widely recognized code repository might be overly complex for a beginner. The proposed iterative workflow allows annotators to verify model predictions and refine the guidelines accordingly. To improve the annotation efficiency, we only ask the annotator to check the samples labeled by the scorer as high-quality and low-quality with high confidence. Besides, there is an automatic validation process in each iteration to ensure the previously annotated samples are correctly classified by the scorer, which is shown as yellow dot lines in the figure. In practice, we took three iterations to finalize our scoring model.

\paragraph{Dependency sorting}
The training context window of InternLM2 has been expanded to 32,000 tokens, allowing the utilization of the entire context of a code repository. The structure of the repository may have been broken in previous data processing steps, such as the filtering of code files by their extensions and deduplication. So we first regroup code files originating from the same repository and perform dependency sorting to establish a sequence for concatenating these files. Therefore, a code repository will be viewed as a single long markdown file with code blocks, which allows the model to learn dependencies across files.

We employ regular expressions to detect the ``import" relations across various programming languages, and we use topological sorting to determine the concatenation order of files. In practice, the arrangement of files may break folder boundaries, resulting in an interleaving sequence of files from multiple sub-folders. Non-code files, such as markdowns and other documentation, are positioned preceding the first code file located in the same sub-folder.

For the corner cases like multiple paths between an ascendant and descendant and loops within the ``import" relation graph, we take the shorter path for the former and use the alphabet order to decide the start point for the latter. A trick in finding ``import" relations is to resolve the batched import, such as ``\_\_init\_\_.py"  or ``\#include xx.h". Those files may import a bunch of unused dependencies, so we apply heuristic rules to refine our detection of ``import" relationships, ensuring that we accurately identify and process these relations at a finer level.

\subsubsection{Long Context Data}
The ability to handle very long contexts ($>32K$ tokens) is an increasingly popular topic in the study of LLMs, broadening and facilitating applications, such as book summarization, supporting long-term conversations, and enabling the handling of tasks involving complex reasoning steps. Pre-training data is one crucial factor for expanding the model's context window. We follow the process of preparing long text pre-training data mentioned in \citet{lv2024longwanjuan}, which includes additional experiments and discussions. The following text only outlines the data preparation employed for InternLM2.

\paragraph{Data Filtering Pipeline}
Our data processing pipeline is designed to filter out low-quality long text data.
It comprises three stages: a) Length selection, a rule-based filter that selects data samples exceeding 32K bytes; b) Statistical filters, which leverage statistical features to identify and remove anomalous data; c) Perplexity filters, which utilize the difference in perplexity to assess the coherence between text segments, filtering out samples with distracting context. Note that all data chosen for the long context training is a subset of the standard pre-training corpus, meaning that the long context data will be learned at least twice during the pre-training.

\paragraph{Statistical Filters}
We employ a variety of lexical and linguistic features to construct our statistical filters. Data samples failing to comply with established rules are excluded from the pre-training corpus. The full list of these filters can be found in \citet{lv2024longwanjuan}. A remarkable filter is the presence of conjunctions and other words that imply discourse structure, such as ``Especially", ``Formally", etc. The overall guidance of designing such filters is to filter out the meaningless data rather than selecting the most high-quality data. Statistical filters are particularly effective with long text data, as the statistical features are far more consistent than those in short text data. For instance, a 20-token text may not yield reliable statistics, but a 32K-token text will have a much clearer statistical feature distribution.

\paragraph{Perplexity filters}
Perplexity is often treated as an estimator for the probability of a text sequence, $P(X)$, and we slightly change its usage to estimate the conditional probability between two text segments $P(S_2 | S_1)$, where $S_1$ is preceding of $S_2$. When the $S_1$ and $S_2$ are strongly correlated, the conditional probability should be higher than estimating the probability of $S_2$ alone, which also means a negative perplexity difference. Conversely, if the probability changed in the reverse direction, meaning that the $S_1$ is a distracting context, it should be removed from the pre-training corpus. Ideally,  adding more context should not negatively impact the predictability of subsequent text. However, we've observed exceptions in cases of improperly joined texts, such as failed HTML parsing, random social media snippets, and other instances stemming from recognition errors in sources with complex layouts. Note that we only filter the data based on the perplexity difference rather than the perplexity itself, and this could largely mitigate the bias introduced by the estimator itself (using which model to compute the perplexity). The bias of the perplexity estimator has been discussed in \citet{DBLP:journals/corr/abs-2402-09739, DBLP:journals/corr/abs-2402-09668}.

\paragraph{Threshold Selection}
Selecting appropriate thresholds is a challenging yet essential part of the data filtering process, and this challenge becomes more severe because we have built many filters. We have two lessons on setting thresholds: 

Tailoring thresholds to each domain rather than seeking a universal solution. For example, the statistical filter for conjunction words would not apply to long code data, which typically does not have any conjunction. Similarly, textbooks, research papers, novels, and patents each exhibit unique characteristics. A universal threshold would likely misclassify a significant amount of data. The same logic applies to setting thresholds across different languages; thus, we adjust thresholds for each domain individually.

Employing a validation set to streamline the process, focusing only on borderline cases. Unlike learning-based feature extractors or scorers, our statistical and perplexity filters yield smooth results within the same domain. It allows us to focus on samples lying near the threshold, simplifying the adjustment of thresholds since we only need to determine whether to lower or raise them. \citet{lv2024longwanjuan} illustrates the data clusters alongside scores from specific filters, demonstrating the interpretability of our proposed filters.

\begin{figure}[h]
    \centering
    \begin{subfigure}[b]{0.35\textwidth}
        \includegraphics[width=\textwidth]{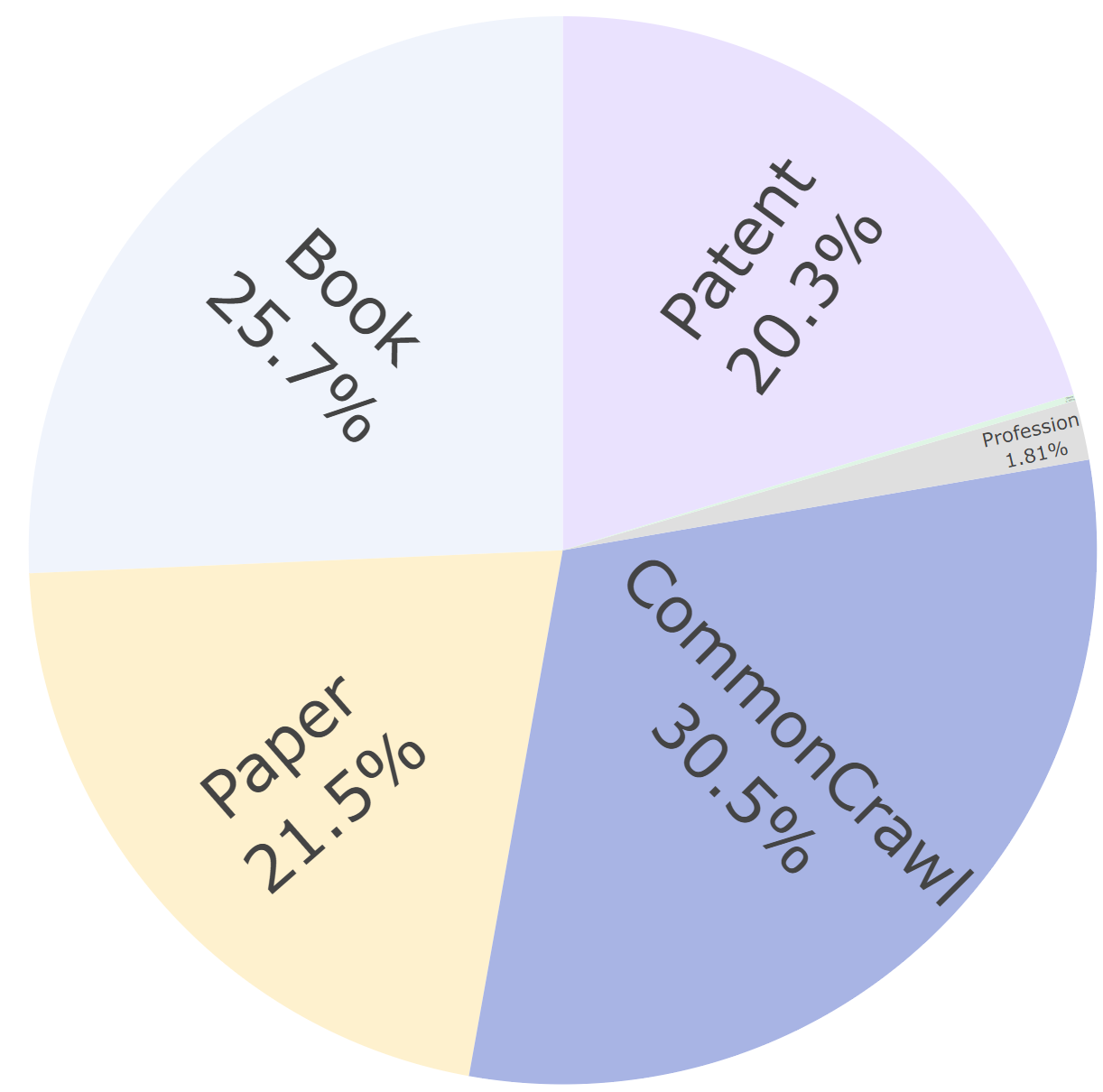}
        \caption{}
        \label{fig:image1}
    \end{subfigure}
    \hfill % 添加一些水平间距
    \begin{subfigure}[b]{0.35\textwidth}
        \includegraphics[width=\textwidth]{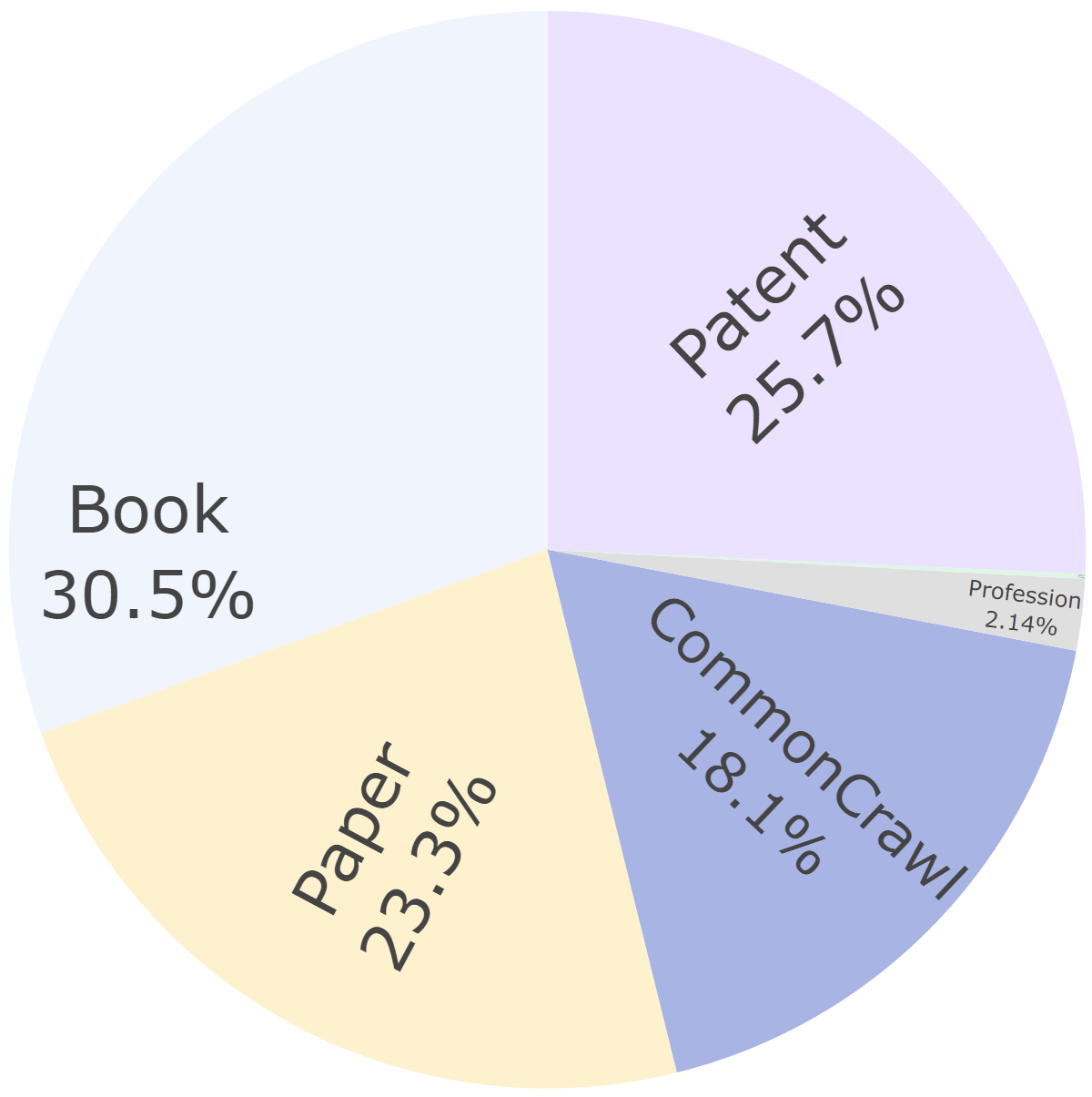}
        \caption{}
        \label{fig:image2}
    \end{subfigure}
    \caption{(a) The statistics of the long text data without filtering. (b) The statistics of the long text data with filtering. }
    \label{fig:long_data_filter}
\end{figure}

Figure~\ref{fig:long_data_filter} shows the data distribution before and after applying all proposed filters. The entire filtering process eliminates a large proportion of Web page data (Common Crawl) and Patent, while most Book and Paper data are persevered.

\subsection{Pre-training Settings}
In this part, we present tokenization and pre-training hyper-parameters. 

\subsubsection{Tokenization}
We have chosen to utilize the tokenization approach of GPT-4 due to its exceptional efficiency in compressing a wide array of textual content. Our primary reference is the cl100k vocabulary~\footnote{\href{https://github.com/openai/tiktoken}{https://github.com/openai/tiktoken}}, which mainly encompasses English and programming language tokens, totaling 100,256 entries, with a minor inclusion of fewer than 3,000 Chinese tokens. To optimize compression rates for InternLM when processing Chinese text, while maintaining the overall vocabulary size under 100,000, we carefully selected the top 60,004 tokens from the cl100k vocabulary and integrated them with 32,397 Chinese tokens. Additionally, we included 147 spare tokens to round out the selection, culminating in a vocabulary size that aligns with a multiple of 256, thereby facilitating efficient training.

\subsubsection{Pre-training Hyper-parameters}

\begin{table}[]
\centering
\begin{tabular}{lcccccc}
\toprule
 Params & $n_{layers}$ & $n_{dim}$ & $n_{kv\_heads}$ & $n_{q\_per\_head}$ & Learning Rate & Batch size \\
\midrule
1.8B &  24&  2048&  8&  2& 3e-4 & 4M \\
7B &  32&  4096&  8&  4& 3e-4 & 4M \\
20B &  48&  6144&  8&  6& 3e-4 & 5M \\
\bottomrule
\end{tabular}
\caption{Hyper-parameters for InternLM2 models.}
\label{tab:training_hyper}
\end{table}

The basic hyper-parameters are listed in Table~\ref{tab:training_hyper}. During training, AdamW~\citep{DBLP:conf/iclr/LoshchilovH19} with $\beta_1=0.9, \beta_2=0.95, \epsilon=1e-8$ and $weight\_decay=0.1$ is used to optimize the model. We use the cosine learning rate decay and the learning rate decays to 10\% of its maximum.

\subsection{Pre-training Phases}
The total number of tokens used for pre-training the 1.8B, 7B, and 20B models ranges from 2.0T to 2.6T, and the pre-training process consists of three distinct phases. In the first phase, we used pre-training corpora with lengths not exceeding 4k. In the second phase, we included 50\% of pre-training corpora with lengths not exceeding 32k. In the third phase, we utilized capability-specific enhancement data. During each phase, we mixed data in English, Chinese, and code.

\subsubsection{4k Context Training}
For approximately 90\% of the training steps, we trained using data with lengths of up to 4096 tokens. If the length of the data exceeded 4096, we would forcibly truncate it and use the remaining part for training as well.

\subsubsection{Long Context Training}
The utilization of extended context windows significantly enhances the performance of Large Language Models (LLMs) in a variety of applications, such as retrieval-augmented generation~\citep{DBLP:journals/corr/abs-2312-10997} and intelligent agents~\citep{DBLP:journals/corr/abs-2309-07864}. Motivated by cutting-edge advancements in training for long context~\citep{DBLP:journals/corr/abs-2308-12950,DBLP:journals/corr/abs-2309-16039,DBLP:journals/corr/abs-2310-05209}, our training process for InternLM2 begins with a 4K context corpus and subsequently transitions to a corpus with 32K context. Instead of only using 32k corpus, 50\% of the data is still shorter than 4096 tokens. This long context training phase accounts for about 9\% of the total steps. 
When accommodated to these longer sequences, we adjusted the Rotary Positional Embedding (RoPE) base from 50,000 to 1,000,000, ensuring more effective positional encoding for long contexts~\citep{DBLP:journals/corr/abs-2310-05209}. Owing to the good scalability of InternEvo~\citep{InternEvo2024} and flash attention~\citep{DBLP:journals/corr/abs-2307-08691}, the training speed only decrease 40\% when changing context window from 4K to 32K.

\subsubsection{Capability Specific Enhancement Training}
Capabilities such as reasoning, mathematical problem-solving, and knowledge memorizing are key abilities expected from large language models.
However, in the pre-training process, high-quality capability-related data is sparsely distributed in the entire corpus, which makes it hard for models to be proficient at these mentioned capabilities. Previous works, such as Qwen~\citep{DBLP:journals/corr/abs-2309-16609}, GLM-130B~\citep{DBLP:conf/iclr/ZengLDWL0YXZXTM23},  Nemotron-4~\citep{parmar2024nemotron}, have tried to incorporate instruction-based or high quality data during the pre-training stage to enhance these abilities.
In InternLM2, we collect an enriched dataset with  a meticulously curated mix of high-quality retrieved data and various types of open-source data from the huggingface
datasets platform~\footnote{\url{https://huggingface.co/datasets}}.
In total, we collect 24 Billion tokens in this dataset,  and details of this corpus are shown in Table~\ref{tab:data-usage}.  
We filter out test set related data and run a contamination test as illustrated in Section~\ref{sec:contamination}. To make the model fit these data well, we employed a smaller learning rate and batch size.

After this enhancement training phase, the InternLM2 model exhibited substantial performance improvements in coding, reasoning, question answering, and examinations, with detailed evaluations results shown in the following evaluation section. 
To aid the research community, we have released checkpoints  before and after the enhancement training, naming them InternLM2-\{size\}-Base and InternLM2-\{size\}, respectively.

\begin{table}[!h]
  \centering
  % \resizebox{.7\textwidth}{!}{
    \begin{tabular}{lccc}
    \toprule
      Category   & Tokens(B)  & Ratio(\%) \\
    \midrule
    \multirow{1}{*}{Retrieved Stem Data}  & 15.97  & 65 \\
    \multirow{1}{*}{Retrieved Special Domain Data}  & 2.17  & 8 \\
    Selected High Quality Data  & 6.26  & 26  \\
    \midrule
   Total &  24.40  & 100  \\
    \bottomrule
    \end{tabular}%
  % }
  \caption{Data details of capability specific enhancement training}\label{tab:data-usage}%
\end{table}%

\section{Alignment}
\label{sec:Alignment}

The pre-training stage empowers LLMs with the foundation abilities and knowledge that are necessary for solving various tasks.
We further fine-tune the LLMs to fully elicit their capabilities and guide LLMs to serve as helpful and harmless AI assistants.
This stage, also commonly referred to as `Alignment', typically contains two phases: supervised fine-tuning (SFT) and reinforcement learning from human feedback (RLHF).
During SFT, we fine-tune the model to follow diverse human instructions by high-quality instruction data (Sec.\ref{subsec:sft}). 
Then we propose \textbf{CO}nditional\textbf{O}n\textbf{L}ine RLHF, which applies a novel conditional reward model that can reconcile different kinds of human preferences (\eg, multi-step reasoning accuracy, helpfulness, harmlessness), and conducts three-round online RLHF to reduce reward hacking (Sec.~\ref{subsec:cool_rlhf}.
In the alignment stage, we keep the long-context capability of LLMs by utilizing long-context pre-training data during SFT and RLHF~\ref{subsec:long_context_finetune}.
We also introduce our practices of improving the tool utilization capability of LLMs~\ref{subsec:tool_utilization}.

\subsection{Supervised Fine-Tuning}\label{subsec:sft}

\begin{figure}[ht]
    \centering
    \includegraphics[width=0.6\linewidth]{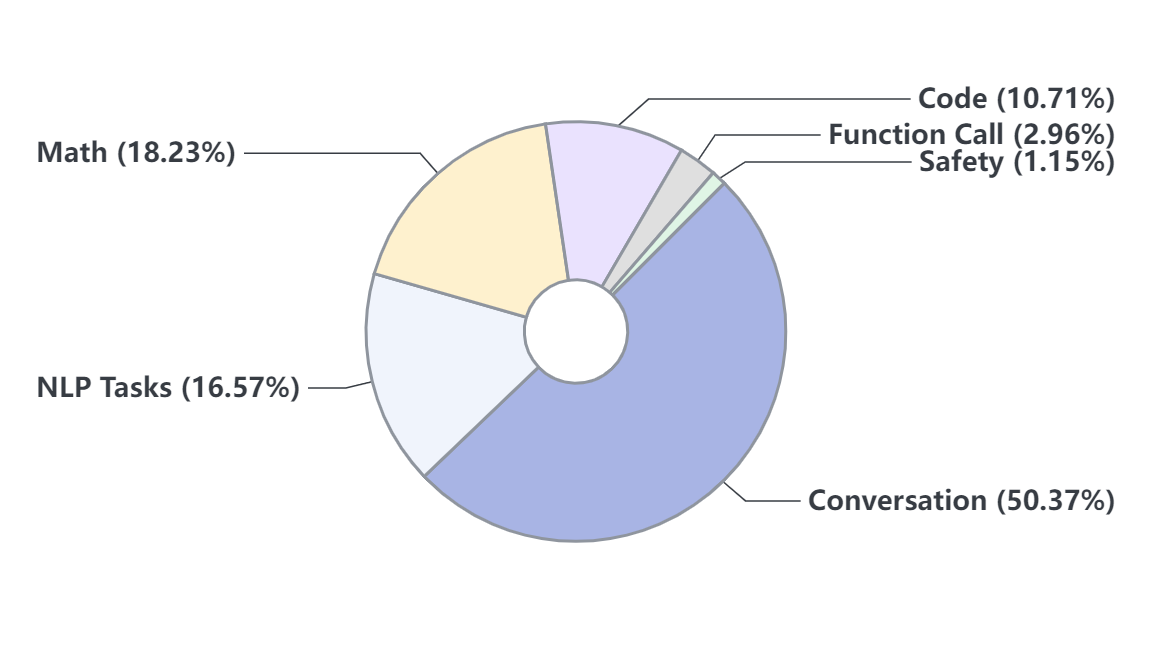}
    \caption{The distribution of SFT data instances.}
    \label{fig:sft_data_distribution}
\end{figure}

In the supervised fine-tuning (SFT) stage, we use a dataset of 10 million instruction data instances, which have been screened to ensure their helpfulness and harmlessness. The dataset encompasses a diverse range of topics, including general conversation, NLP tasks, mathematical problems, code generation and function calls, etc. Figure \ref{fig:sft_data_distribution} shows the detailed distribution of SFT data topics. To facilitate a versatile representation of such various tasks, we transform the data samples into the ChatML~\citep{ChatML} format. Both the 7B and 20B models undergo training for one epoch using the AdamW optimizer with an initial learning rate of 4e-5.

\subsection{COOL Reinforcement Learning from Human Feedback}\label{subsec:cool_rlhf}

Reinforcement Learning from Human Feedback (RLHF)~\citep{DBLP:conf/nips/ChristianoLBMLA17, DBLP:conf/nips/Ouyang0JAWMZASR22} is an innovative approach within the realm of large language models. By incorporating human feedback, RLHF creates reward models that serve as proxies for human preferences, thereby providing reward signals for LLMs to learn through the use of Proximal Policy Optimization (PPO)~\citep{DBLP:journals/corr/SchulmanWDRK17}. This methodology enables models to better understand and execute tasks that are difficult to define through traditional methods.

Despite the achievements of RLHF, there are still some issues in its practical application. The first is the preference conflicts. For example, in developing a dialogue system, we expect it to provide useful information (helpful) while not producing harmful or inappropriate content (harmless). However, these two preferences often cannot be satisfied simultaneously in practice, as providing useful information might involve sensitive or high-risk content in some cases. Existing RLHF methods~\citep{DBLP:journals/corr/abs-2307-09288, dai2023safe, wu2023fine} usually rely on multiple preference models for scoring, which also introduces more models in the training pipeline thus increases computational cost and slows down training speed.
Second, RLHF faces the issue of reward hacking, especially when the policy becomes more powerful with the scale increasing~\citep{Manheim_Garrabrant_2018, Gao_Schulman_Hilton_2022}, where the model might learn to ``cheat" the reward system by shortcuts to obtain high scores, rather than truly learning the expected behavior. This leads the model to maximize rewards in unintended ways, significantly affecting the effectiveness and reliability of LLMs.

To address these issues, we propose \textbf{Co}nditional \textbf{O}n\textbf{L}ine RLHF (COOL RLHF). COOL RLHF first introduces a Conditional Reward mechanism to reconcile diverse preferences, which allows the reward model to dynamically allocate its attention to various preferences based on specific conditions, thereby optimally integrating multiple preferences. Furthermore, COOL RLHF adopts a multi-round Online RLHF strategy to enable the LLM to promptly adapt to new human feedback, mitigating the occurrence of reward hacking. 

\subsubsection{Conditional Reward Model}

\begin{figure}[ht]
    \centering
    \includegraphics[width=0.8\linewidth]{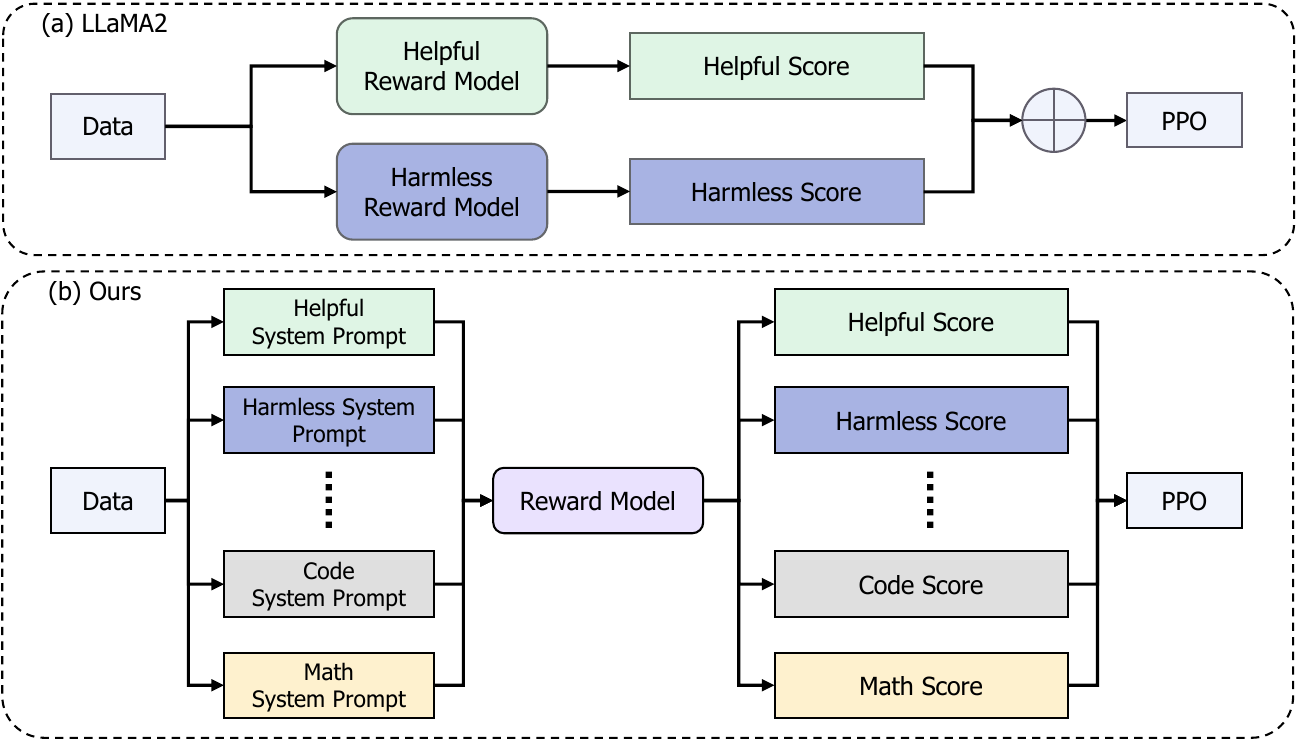}
    \caption{Architecture of the Conditional Reward Model. (a) LLaMA2 employs distinct reward models to address the issue of preference conflicts. (b) The proposed conditional reward model (ours) utilizes conditional system prompts to reconcile preference data across various domains, enabling the modeling of multiple preferences using a single reward model.}
    \label{fig:conditional_reward}
\end{figure}

The Conditional Reward Model represents an innovative solution to the challenges inherent in the previous preference modeling of RLHF methods. Unlike conventional approaches that typically rely on multiple preference models to address preference conflicts across different domains (Figure~\ref{fig:conditional_reward}(a)), the Conditional Reward Model incorporates different system prompts for different kinds of preferences to effectively model a variety of preferences within a singular reward model. 

Specifically, as depicted in Figure~\ref{fig:conditional_reward}(b), the Conditional Reward Model employs different system prompts to seamlessly blend data from various fields. Since the reward model is initialized from a SFT model, which already learned to follow diverse human instructions, we also let the reward model follow different system prompts to adapt to diverse preferences of different scenarios. In the Conditional Reward Model, system prompts are not simply a component of its input; they are also a crucial tool for directing the reward score in alignment with specific preferences in varied scenarios.
Such an integration facilitates the management of contradictory and complex human preferences within a unified reward model without sacrificing accuracy. 

\paragraph{Data Composition}

The training process of the Conditional Reward Model involves an extensive dataset, encompassing various fields such as dialogue, article writing, poetry, summarization, coding, mathematics, and formatted output, with up to 2.4 million binarized preference pairs. This comprehensive dataset ensures the model's broad adaptability and enhances its capability to undertake reinforcement learning in a broader and more complex array of situations. Thus, by employing the conditional system prompt method, the reward model can respond to complex human requirements, providing a more nuanced control over the reward scores during the PPO phase.

\paragraph{Loss Function}

Furthermore, to reduce the influence of the imbalance between easy and difficult samples in the dataset, we modify the original ranking loss function~\citep{DBLP:conf/icml/BurgesSRLDHH05} inspired by Focal Loss~\citep{DBLP:conf/iccv/LinGGHD17}. 
We add a difficulty decay coefficient to the ranking loss, making the loss value larger for difficult samples and smaller for easy ones, preventing overfitting on a large number of easy samples. The focal ranking loss is formulated as

\begin{equation}
    L_{ranking}= - (1 - 2 \times \max (0, P_{i,j}-\frac{1}{2}))^{\gamma} \log(P_{i,j})),
\end{equation}

where $P_{i,j} = \sigmoid(r_{i} - r_{j})$ represents the probability that $reward_{i}$ is greater than $reward_{j}$. The difficulty decay coefficient only takes effect when the model correctly predicts the preference of the training sample, i.e., $P_{i,j}>0.5$, otherwise it equals to 1. The term $\gamma$ represents a hyper-parameter that is instrumental in modulating the difficulty decay ratio. Here we set it to 2 by default. Concurrently, to ensure the stability and consistency of the output scores from the reward model across different training, we introduce a logarithmic barrier penalty to the reward scores to confine the score distribution within a range of -5 to 5, defined as

\begin{equation}
    L_{penalty}= - (\log(x + 5) + \log(5 - x)).
\end{equation}

This constraint is critical as it obviates the need to modify additional reward-related hyper-parameters in the PPO stage, which could potentially arise due to variations of the reward score distribution in different reward models. 
Overall, the loss function of the reward model is

\begin{equation}
    L=L_{ranking} + \lambda L_{penalty}.
\end{equation}

The parameter $\lambda$ is a weighting coefficient that balances the contribution of $L_{ranking}$ and $L_{penalty}$. We set it to a default value of 0.02 based on our observation in the preliminary experimental results. These enhancements improve the robustness and consistency of the reward model, particularly in the context of datasets characterized by an imbalance between easy and difficult samples.

\paragraph{Training Details}

In our experiment, we align the sizes of the reward models with those of the actor models used in PPO. Following the methods described in InstructGPT\citep{DBLP:conf/nips/Ouyang0JAWMZASR22}, we initialize the reward models using the SFT model weights, modifying the output layer to a one-dimensional linear mapping layer, which was randomly initialized. Our batch construction strategy focuses on fixing the total length of preference data at 16384 tokens per batch, rather than limiting the number of preference pairs, to avoid training inefficiencies due to data padding. The maximum context length is set at 8192. A special token is appended to each sequence's end, with its output value used as the reward score. We adapt AdamW as the optimizer. The learning rate follows a cosine annealing schedule, decreasing from 1e-5 to 5e-6 and weight decay is set to 0.01. To prevent overfitting, the models are trained for one epoch.

\subsubsection{Online RLHF}
After obtaining a conditional reward model, we conduct Proximal Policy Optimization (PPO) to align the LLMs to the human preferences modeled by the reward model~\cite{DBLP:conf/nips/Ouyang0JAWMZASR22}.
To address the challenge of reward hacking in the PPO stage, we introduce an Online RLHF approach, divided into two distinct pathways: a Fast Path for immediate, targeted improvements and a Slow Path for long-term, comprehensive refinement of the reward model. The Fast and Slow Paths are complementary to provide an adaptive framework for mitigating reward hacking and enhancing the performance and reliability of LLMs trained with human feedback.

\paragraph{Fast Path} 
The fast path in Online RLHF focuses on quickly identifying and rectifying reward hacking incidents through targeted patches to improve the reliability of the reward model. As the PPO training goes by, the LLMs are encouraged to gravitate towards high-reward regions, which usually expose more reward hacking scenarios that can be easily detected.
After identifying the hacking pattern after each round of RLHF, we construct preference pairs that highlight these patterns by comparing responses generated by early and late-stage PPO models in the current round. Incorporating 20 to 100 such preference pairs into the training process is sufficient to prevent the reward model from the corresponding hacking pattern significantly. This process allows for a swift fix of the reward model to the emerging hacking behaviors, enhancing the reward model's reliability and adherence to desired outcomes.

\paragraph{Slow Path} 
In comparison to the fast path that focuses on fixing reward hacking, the slow path aims at a general improvement of the reward model's upper bound, especially the reliability and robustness of the reward model at the high-reward regions, by covering the LLMs responses from the most recent and capable models, following previous work~\citep{bai2022training}. To achieve this, responses generated by models at various stages of training (including the SFT model, early-stage PPO model, and late-stage PPO model) are used to form pairwise comparisons. These pairs are then presented to professional human annotators to label their preferences.

Such a process provides a more nuanced and thorough refinement of the reward model but requires extensive human annotation time. To improve the efficiency of online RLHF, we only use the accumulated human preferences of all previous models at the launch time of our experiments (\textit{i.e.}, which may not include the preferences of responses produced by models at the current round due to the time cost of human labeling). By continuously updating the model based on human feedback, the Slow Path ensures that the reward model evolves in tandem with the complexities and subtleties of human preferences.

\paragraph{Implementation}
In our implementation of Online RLHF, we conducted three rounds of refinement. In these cycles, we gathered thousands of preference patches and online preference data in the Fast Path to update the reward model, and use all the existing human preference data of responses from previous models.
Each round of Online RLHF provided valuable insights, allowing us to dynamically adjust and refine the reward model, thereby enhancing the overall performance and reliability of language models trained with human feedback.

\subsubsection{PPO Training Details}

During the RL alignment phase, we adopted the standard PPO (Proximal Policy Optimization) algorithm and made several adaptions to it, ensuring a more stable training process. The framework involves four models: the actor model, critic model, reference model, and reward model. During training, the latter 2 models are frozen, and only the former 2 models are actively trained. Notably, all these models are of the same size, ensuring consistency in their capacity to process and generate data. We iterate through about 200k diverse queries in about 400 iterations and choose the best checkpoints on the validation sets for release.

\paragraph{Model Initialization}

\begin{figure}[ht]
    \centering
    \includegraphics[width=0.99\linewidth]{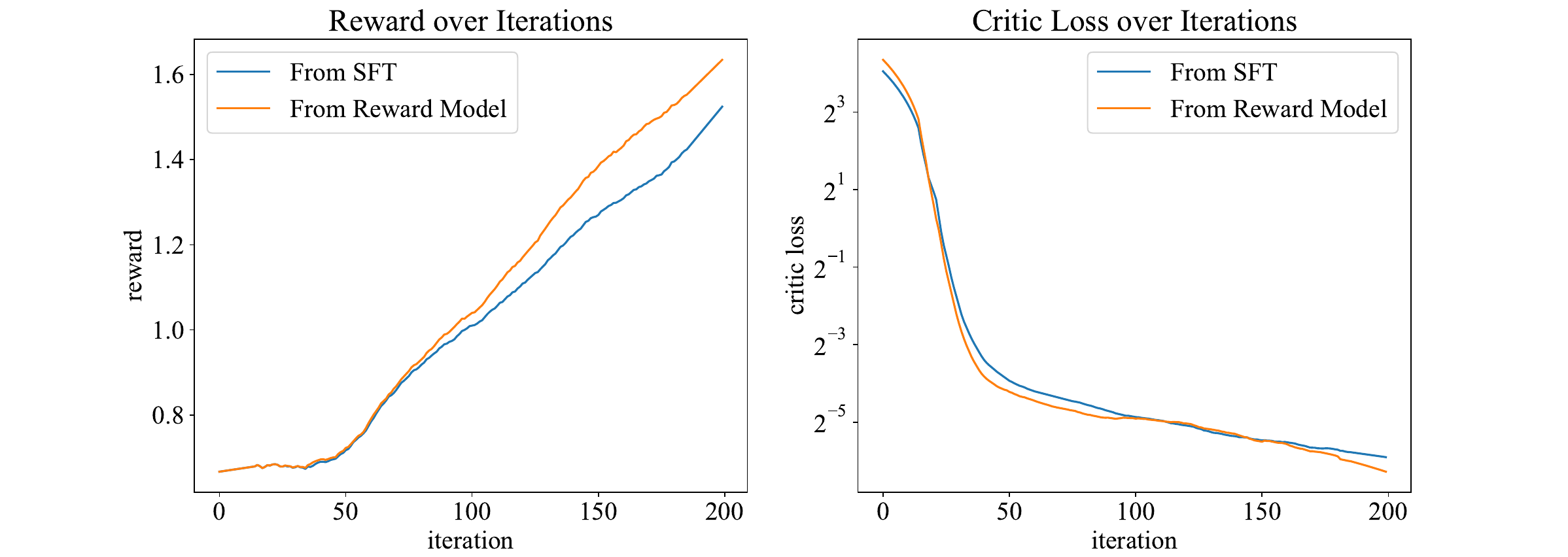}
    \caption{Ablation study on critic model initialization. Note that we used logarithmic coordinates in critic loss due to its large dynamic range.}
    \label{fig:ablation_critic_initialization}
\end{figure}

As is common practice, we initialize the reference model and the actor model from the SFT model weights. The critic model is initialized from the reward model (excluding the linear head) and undergoes a 50-iteration pre-training phase, during which the actor model is frozen. This phase is critical for stabilizing the value estimation in early training, thus preventing the potentially detrimental effects of unstable values. We conducted ablation studies comparing the initialization of the critic model from the reward model versus the SFT model, as show in Figure~\ref{fig:ablation_critic_initialization}. Our results show that the critic model initialized from the reward model experiences larger losses during the first few iterations of PPO training, but after approximately 20 iterations, it consistently exhibits lower losses and leads to higher rewards for the actor model. We hypothesize that the higher loss observed during the initial stage may reveal fundamental differences between the tasks of reward modeling and critic modeling. The subsequent lower loss could be attributed to a more consistent internal understanding of the world knowledge and a better grasp of the assessment principles.

\paragraph{Conditional Reward}

\begin{figure}
    \centering
    \includegraphics[width=0.8\linewidth]{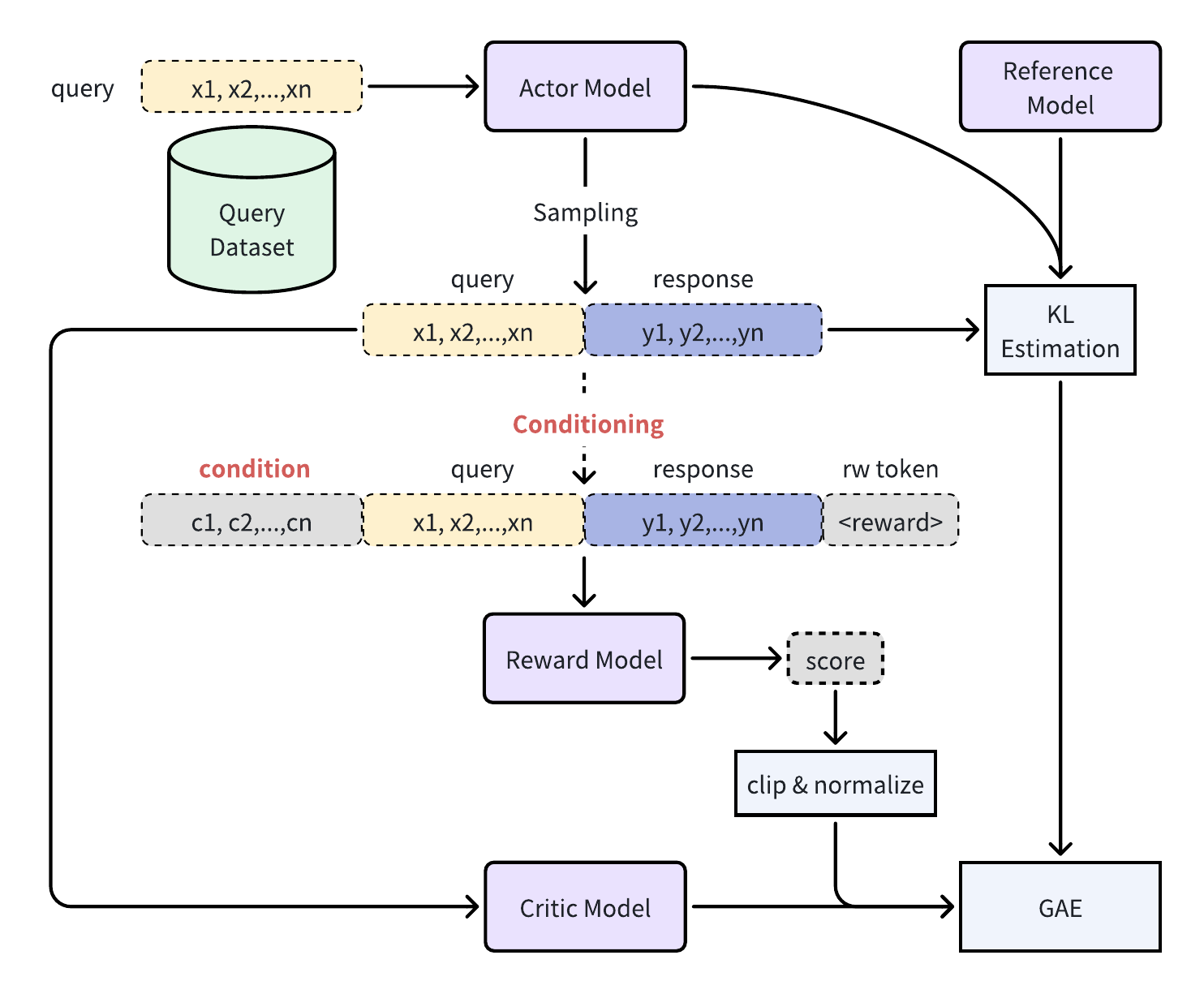}
    \caption{Illustration of conditional PPO training. An appropriate conditional system prompt is added to query \& response before reward model assesment. Note that this system prompt is agnostic to the other three models.}
    \label{fig:conditional_ppo}
\end{figure}

As discussed earlier, our reward models are trained to adapt to various conditions. Consequently, for queries from different domains, we prepend appropriate conditional system prompts to every sampled response before calculating reward scores, as illustrated in Figure~\ref{fig:conditional_ppo}. This practice ensures that the model's responses are contextually aligned with the varied demands of different domains.

\paragraph{Pre-train Gradients}

To mitigate the risk of catastrophic forgetting during the PPO phase, we incorporate a pre-train loss, following the InstructGPT methodology. The coefficient for the pre-train loss is set at 0.5, and the volume of the pre-train data is approximately 50\% of that for PPO training. This addition helps to preserve the knowledge acquired during the initial training stages, ensuring that the model retains its foundational abilities and knowledge base while adapting to new feedback and learning through PPO.

\paragraph{Hyperparameters}

We set the KL divergence coefficient to 0.01. The learning rates for the actor model and the critic model are set to 1e-6 and 5e-6, respectively. We found that a larger $\lambda$ value for PPO leads to higher rewards in our case, so we set it to 0.99. We adopted a slightly conservative sampling strategy, with $top\_p=0.9$, to strike a balance between sampling diversity and convergence speed. Unlike some conventional approaches, we do not apply value loss clipping or advantage normalization. Despite extensive RL tricks, the training remains remarkably stable, partially due to our meticulous Online RLHF efforts.

\subsection{Long-Context Finetuning}\label{subsec:long_context_finetune}
To preserve the long-context capability of LLMs after fine-tuning, we keep using the long-context pre-training data in SFT and RLHF, inspired by previous work that adopts long-context pre-training corpus in SFT~\citep{DBLP:journals/corr/abs-2309-16039}. Specifically, we utilize two types of data: one comprises long-context data from books, while the other is long-context data obtained from GitHub repositories and concatenated through specific paradigms, described below.

\begin{figure}[h!]
    \centering
    \includegraphics[width=0.8\linewidth]{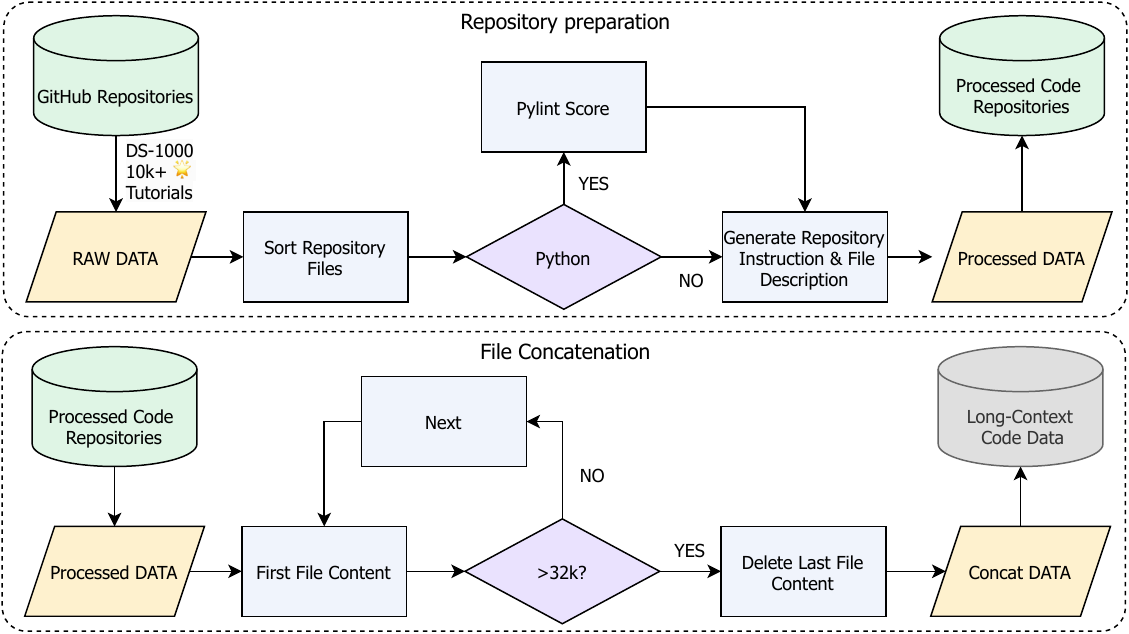}
    \caption{Illustration of the process to obtain long-context code data.}
    \label{fig:long_context_code_data}
\end{figure}

To enhance the data analysis capabilities of InternLM2, we choose the code repositories used in DS-1000~\citep{ds-1000} as the core repositories, which include Pandas, Numpy, Tensorflow, Scipy, Scikit-learn, PyTorch, and Matplotlib. We then search for repositories on GitHub with over 10,000 stars that referenced these core repositories and conducted the same filtering and data cleaning process as pre-training. 
For each repository, we initially sort the acquired raw data using a depth-first approach, while simultaneously generating the required prompts that briefly describe the file content, as shown in Figure~\ref{fig:long_context_code_data}. Subsequently, we concatenate the processed data in sequence until reaching a length of 32k.
The experimental results show that long-context code data improves not only the long-context capability of LLMs but also the code capabilities.

\subsection{Tool-Augmented LLMs}\label{subsec:tool_utilization}
\paragraph{General Tool Calling} We adopt a modified version of the ChatML format to enable general tool calling by introducing the ``environment'' role. Such a modification shares the same format in chat scenarios but provides more clear signal to the model when adopting agents. Additionally, we define two specific keywords to support the diverse purpose of AI agents, \textit{i.e.,} code interpreter (\texttt{<|interpreter|>}) and external plugins (\texttt{<|plugin|>}). This enables us to adopt a unified streaming format that can handle various types of plugin extensions and AI environments while being compatible with general chat. Figure \ref{fig:func_call} shows a concrete example of streaming chat format. To fully elicit the agent ability of InternLM2, we align the agent corpus to the chat domain and disentangle it along the basic capabilities of the language model for fine-grained training, as described in Agent-FLAN~\citep{chen2024agent}.

\paragraph{Code Interpreter}
We also enhance the ability of InternLM2-Chat to solve the math problems by code interpreters, by treating the Python code interpreter as a special tool using the same schema described in Tool Learning. We adopt the reasoning interleaved with coding (RICO) strategy and construct the data in an iterative hard example mining manner, described in InternLM-Math~\citep{internlmmath}. 

\section{Evaluation and Analysis}

\subsection{Overview}
 In this section, we provide a comprehensive evaluation and analysis of the language model’s performance across various domains and tasks. The evaluation is structured into two main categories: 
 \textit{(a) downstream tasks} and \textit{(b) alignment}. For each category, we further break down the evaluation into specific subtasks to provide a detailed understanding of the model’s strengths and weaknesses. Lastly, we discuss the potential issue of data contamination in language models and its impact on model performance and reliability. All evaluations are performed using OpenCompass~\citep{2023opencompass} unless explicitly specified otherwise\footnote{\href{https://github.com/open-compass/opencompass}{https://github.com/open-compass/opencompass}}.

\subsection{Performance on Downstream Tasks}
We start by detailing evaluation protocols and performance metrics for multiple NLP tasks. We introduce datasets, explain our experimental setup, and then present results with in-depth analysis, comparing to state-of-the-art methods to showcase the effectiveness of our model. The performance assessment will be dissected through six key dimensions: \textit{(1) comprehensive examinations, (2) language and knowledge, (3) reasoning and mathematics, (4) multiple programming language 
coding, (5) long-context modeling, (6) tool utilization}.

\subsubsection{Comprehensive Examination}

We conducted benchmarks on a series of exam-related datasets, which include:

\noindent

\noindent \textbf{MMLU}~\citep{hendrycks2020measuring}: A multiple-choice question dataset containing 57 subtasks, covering topics in humanities, social science, STEM, and others. We report 5-shot results.

\noindent \textbf{CMMLU}~\citep{li2023cmmlu}: A multiple-choice question dataset specific to China, containing 67 subtasks. In addition to humanities, social science, STEM, and others, it includes many Chinese-specific tasks.  We report 5-shot results.

\noindent \textbf{C-Eval}~\citep{huang2023ceval}: A multiple-choice question dataset containing 52 subtasks and 4 difficulty levels, covering topics in humanities, social science, STEM, and others.  We report 5-shot results.

\noindent \textbf{AGIEval}~\citep{zhong2023agieval}: A human-centric benchmark that includes both multiple-choice and open-ended questions. The questions are from 20 official, public, and high-standard admission and qualification exams intended for general human test-takers and report 0-shot results.

\noindent \textbf{GAOKAO-Bench}~\citep{Zhang2023EvaluatingTP}: A dataset containing the Chinese college entrance examination (Gaokao) from 2010 to 2022, including both subjective and objective questions. We only evaluated the dataset of objective questions and report 0-shot results.

\input{sections/evaluation/examination_table}

\paragraph{Evaluation Results}

\noindent

We report the result of base models in Table~\ref{tab:exam_base_results} and chat models in Table~\ref{tab:exam_chat_results}.

For the base model, the InternLM2 series performs well among models with similar number of parameters. On 7B and 20B, InternLM2 shows a significant increase compared to InternLM2-Base, which proves that pre-training on general domain data and domain-enhanced corpus has advantages for comprehensive examination. For the AGIEval and GAOKAO tasks, which are collected from exams designed specifically for human, InternLM2 shows a greater increase compared to InternLM2-Base relative to other datasets.

For the chat model, the InternLM2 series also excels among models with similar number of parameters. By comparing the InternLM2-Chat-7B-SFT and InternLM2-Chat-7B models, it can be seen that COOL RLHF has little impact on comprehensive examination performance.

\subsubsection{Language and Knowledge}
\input{sections/evaluation/language_table_1}

\noindent \textbf{TriviaQA}~\citep{joshi2017triviaqa}: A dataset containing both reading comprehension and open-domain QA. On average, each question has 6 possible answers. We utilized the open-domain QA subset of the data and report 0-shot results.

\noindent \textbf{NaturalQuestions}~\citep{naturalquestion}: A dataset of QA where the questions come from users and the answers are verified by experts. We report 0-shot results.

\noindent \textbf{C3}~\citep{sun2019investigating}: A free-form multiple-choice Chinese machine reading comprehension dataset. We report 0-shot results.

\noindent \textbf{RACE}~\citep{lai-etal-2017-race}: A reading comprehension dataset that includes English reading comprehension exam questions for Chinese middle and high school students aged 12 to 18. We use the subset for high school students and report 0-shot results.

\noindent \textbf{FLORES}~\citep{nllb2022}: A translation dataset extracted from Wikipedia, covering 101 languages. We evaluated the translation results from English to the other 100 languages and vice versa. For each pair of translation tasks, we selecte 100 samples and evaluate with BLEU. We report 8-shot results.

\paragraph{Evaluation Results}

\noindent

We report the result of base models in Table~\ref{tab:language_base_results} and chat models in Table~\ref{tab:language_chat_results}. Thanks to its rigorous and high-quality training corpus, InternLM2 has demonstrated a remarkable competitive edge in tasks that involve language understanding and knowledge application. Consequently, it emerges as an excellent choice for a multitude of real-world applications where the reliance on robust language comprehension and extensive knowledge is paramount.

\subsubsection{Reasoning and Mathematics}

In this section, we will primarily validate the performance of InternLM2 in reasoning and mathematics, focusing on the following two parts of the test sets:

\textbf{Reasoning Datasets:}

\noindent

\noindent $\bullet$  \textbf{WinoGrande}~\citep{DBLP:conf/aaai/SakaguchiBBC20}: A commonsense reasoning dataset containing 44,000 multiple-choice questions with two options each. It requires the model to choose the appropriate entity word for the pronoun in the descriptive text based on the scenario.

\noindent $\bullet$ \textbf{HellaSwag}~\citep{DBLP:conf/acl/ZellersHBFC19}: A challenging dataset for evaluating commonsense natural language inference, consisting of 70,000 multiple-choice questions. Each question presents a scenario and four possible outcomes, asking to select the most reasonable conclusion.

\noindent $\bullet$  \textbf{BigBench Hard (BBH)}~\citep{DBLP:conf/acl/SuzgunSSGTCCLCZ23}: A test collection for large language models, BBH extracts 23 challenging tasks from BIG-Bench, where contemporary language models had not surpassed human performance at the time.

\textbf{Mathematics Datasets:}

\noindent

\noindent $\bullet$ \textbf{GSM8K-Test}~\citep{DBLP:journals/corr/abs-2110-14168}: A dataset containing approximately 1,300 elementary-level situational math problems. The solution to these problems involves 2 to 8 steps, primarily using basic arithmetic operations (addition, subtraction, multiplication, and division) to perform a series of basic calculations to reach the final answer.

\noindent $\bullet$ \textbf{MATH}~\citep{DBLP:conf/nips/HendrycksBKABTS21}: A dataset of 12,500 challenging high school-level competition math problems, covering multiple areas from algebra to calculus. Each problem includes a complete step-by-step solution.

\noindent $\bullet$ \textbf{TheoremQA}~\citep{DBLP:conf/emnlp/ChenYKLWMXWX23}: A STEM theorem-driven question and answer dataset containing 800 QA pairs, covering over 350 theorems in mathematics, EE\&CS, physics, and finance. It tests the limitations of large language models in applying theorems to solve challenging university-level problems.

\noindent $\bullet$ \textbf{MathBench}~\citep{Anonymousmathbench}: MathBench comprises 3709 questions with multiple stages of progressively increasing challenges. Each stage encompasses bilingual theoretical and application-oriented questions, with each question precisely tagged with a three-level label to indicate its fine-grained knowledge point.

\begin{table}[!htb]
    \captionsetup{justification=centering}

    \begin{minipage}{.5\linewidth}
      \centering
        
        \resizebox{\linewidth}{!}{
\begin{tabular}{lccc}
\hline
\textbf{Models} & \textbf{WinoGrande} & \textbf{HellaSwag} & \textbf{BBH} \\
& \textit{0-shot} & \textit{0-shot} & \textit{0-shot} \\
\hline
\multicolumn{4}{c}{\greencolor{\textit{$\triangle$ $\sim7B Models$}}} \\
ChatGLM3-6B-Base & 51.4 & 76.5 & \underline{66.2} \\
Baichuan2-7B-Base & 68.7 & 70.2 & 41.9 \\
Llama2-7B & 69.5 & 74.0 & 39.3 \\
Mistral-7B-v0.1 &  75.3 & 78.9 & 56.7 \\
Qwen-7B & 68.6 & 75.0 & 45.2 \\
\midrule
InternLM2-7B-Base & 74.6 & 76.3 & 55.5 \\
InternLM2-7B & \underline{84.7} & \underline{79.3} & 65.0 \\
\multicolumn{4}{c}{\bluecolor{\textit{$\heartsuit$ $\sim20B Models$}}} \\
Llama2-13B & 72.3 & 77.5 & 47.2 \\
Baichuan2-13B-Base & 70.0 & 73.7 & 48.9 \\
Qwen-14B & 67.5 & 80.3 & 53.7 \\
Mixtral-8x7B-v0.1 & 77.3 & 81.9 & 68.8 \\
\midrule
InternLM2-20B & \textbf{85.2} & \textbf{81.6} & \textbf{72.1} \\
InternLM2-20B-Base & 76.2 & 78.1 & 62.0 \\
\hline
\end{tabular}
        }
        \caption*{\small{Base Model Results.}}
    \end{minipage}%
    \begin{minipage}{.5\linewidth}
      \centering
        
        \resizebox{\linewidth}{!}{
\begin{tabular}{lccc}
\hline
\textbf{Models} & \textbf{WinoGrande} & \textbf{HellaSwag} & \textbf{BBH} \\
& \textit{0-shot} & \textit{0-shot} & \textit{0-shot} \\
\hline
\multicolumn{4}{c}{\purplecolor{\textit{\ding{72}API Models}}} \\
GPT-3.5 & 68.7 & 70.2 & 41.9 \\
\multicolumn{4}{c}{\greencolor{\textit{$\triangle$ $\sim7B Models$}}} \\
ChatGLM3-6B & 61.7 & 73.1 & 56.0 \\
Llama2-7B-Chat & 51.5 & 49.5 & 41.4 \\
Baichuan2-7B-Chat & 49.9 & 36.4 & 35.9 \\
Mistral-7B-Instruct-v0.2 & 50.8 & 64.1 & 46.4 \\
Qwen-7B-Chat & 54.2 & 61.9 & 45.5 \\
\midrule
InternLM2-Chat-7B-SFT & 65.8 & \underline{83.5} & 60.9 \\
InternLM2-Chat-7B & \underline{65.8} & 83.0 & \underline{61.2} \\
\multicolumn{4}{c}{\bluecolor{\textit{$\heartsuit$ $\sim20B Models$}}} \\
Llama2-13B-Chat & 50.8 & 57.0 & 49.7 \\
Baichuan2-13B-Chat & 50.9 & 34.4 & 42.5 \\
Mixtral-8x7B-Instruct-v0.1 & 60.9 & 80.3 & 57.3 \\
Qwen-14B-Chat & 55.7 & 79.2 & 55.8 \\
\midrule
InternLM2-Chat-20B-SFT & \textbf{75.0} & \textbf{85.9} & 66.9 \\
InternLM2-Chat-20B & 74.8 & 85.8 & \textbf{68.3} \\
\hline
\end{tabular}
        }
        \caption*{\small{Chat Model Results.}}
    \end{minipage}
        \caption{\textbf{Comparison of Reasoning Tasks}}
    \label{tab:reasoning_result}
\end{table}

In the evaluation of reasoning and mathematics problems, for multiple-choice questions, we predominantly use the zero-shot approach. For open-ended questions, such as those in GSM8k, MATH, and the open-ended section of MathBench, we primarily employ a few-shot methodology to enhance the model's ability to follow instructions, which facilitates the extraction of answers. To ensure consistency in evaluation results, for the base models, we utilize perplexity (PPL) evaluation as the main method for assessing multiple-choice questions.

\paragraph{Reasoning}
Reasoning ability reflects a model's capacity to understand, process, and manipulate abstract concepts, which is essential for tasks involving complex problem-solving and decision-making. We separately compare and demonstrate the performance of InternLM2 in \Cref{tab:reasoning_result} from two aspects: Base and Chat.

\paragraph{Evaluation Results}
\quad

\textbf{Base Model:}
For models around 7B, InternLM2-7B demonstrates superior performance on most datasets outside of BBH. Notably, its performance on WinoGrande (84.7) significantly surpasses that of Mistral-7B-v0.1 (75.3) by 9.4 points, ranking second only to its series counterpart InternLM2-20B (85.2) by less than one point. This showcases the strong commonsense reasoning capabilities of InternLM2-7B. Among all tested base models, InternLM2-20B achieves the best overall performance. Compared to InternLM2-20B-Base, the inclusion of domain-enhanced knowledge has led to noticeable improvements in commonsense reasoning, with InternLM2-20B showing an average increase of 10.4\% across the tested reasoning datasets.

\textbf{Chat Model:}
In terms of reasoning within chat models, InternLM2 continues to lead, both in the 7B phase and the 13$\sim$20B phase. RL models and SFT models exhibit similar effects on a significantly advanced basis of reasoning capabilities. Among these, the 7B parameter models even outperform most of the 13$\sim$20B models on three test sets, such as Mixtral-8x7B-Instruct-v0.1 and Qwen-14B-Chat, and InternLM2-Chat-20B significantly outperforms GPT-3.5 across various aspects of reasoning test sets, such as in situational reasoning on HellaSwag (\(\uparrow 15.6\)) and challenging comprehensive reasoning on BBH (\(\uparrow 26.4\)), demonstrating InternLM2's exceptional reasoning abilities in smaller-scale models.

\paragraph{Mathematics}

\input{sections/evaluation/math_table_1}

\input{sections/evaluation/math_table_2}

Mathematical proficiency is an integral element of a model's cognitive and computational faculties. In~\Cref{tab:base_math_result}, we present the performance of the Base model across multiple math assessment sets of varying difficulty. At the 7B parameter scale, InternLM2-7B takes the leading positions. Specifically, in the basic arithmetic GSM8k set, InternLM2-7B (70.8) significantly outperforms other models, surpassing ChatGLM3-6B-Base (60.7) by 10.1 percentage points and nearly doubling the performance relative to InternLM2-7B-Base (36.0), demonstrating the powerful basic arithmetic capability of InternLM2-7B after incorporating domain-enhanced data. For more complex problem-solving tasks such as MATH and theorem proving in TheoremQA, InternLM2-7B, despite being slightly behind in MATH, achieves a score (10.5) that even surpassed the larger Qwen-14B-Base (10.4), indicating that InternLM2-7B excels not only in computational tasks but also in complex theorem proving. In the Bilingual math dataset MathBench, InternLM2 shows excellent performance both in English and Chinese.

In the 13$\sim$20B parameter scale phase, InternLM2-20B outperforms all tested Base models in basic arithmetic and theorem proving, while Qwen-14B-Base demonstrates outstanding results in problem-solving tasks and both English and Chinese tests of MathBench.

\textbf{Chat}
For Chat models in \Cref{tab:chat_math_result}, InternLM2-Chat demonstrates the best performance in basic arithmetic GSM8K, complex applied mathematics MATH, and theoretical problems TheoremQA at both the 7B and 20B parameter scales. Notably, InternLM2-Chat-20B, which underwent COOL RLHF training, leads comprehensively across all metrics, outperforming the API model GPT-3.5 and the MoE model Mixtral-8x7B-Instruct-v0.1. In the bilingual tests of MathBench, InternLM2-Chat-20B also shows excellent performance.

\subsubsection{Coding}

\paragraph{Python Coding Tasks}

\noindent

\noindent $\bullet$  \textbf{HumanEval} 
HumanEval~\citep{chen2021evaluating} is a widely recognized dataset that serves as a benchmark for evaluating the performance of code generation models. It consists of 164 carefully crafted programming tasks, each composed of a Python function and an accompanying docstring to provide context and specifications. This dataset, featuring human-written code, plays a pivotal role in assessing the capabilities of Large Language Models (LLMs) when generating or completing programs.

\noindent $\bullet$ \textbf{MBPP}
MBPP~\citep{austin2021program} consists of 974 programming tasks that are solvable by entry-level programmers. These tasks range from simple numeric manipulations to more complex problems requiring external knowledge, such as defining specific integer sequences. We use the santinized version of MBPP, which only includes a subset of the data has been hand-verified by the authors.

\paragraph{Multiple Programming Language Coding Tasks}

\noindent

\noindent $\bullet$ \textbf{HumanEval-X} HumanEval-X~\citep{zheng2023codegeex}  dataset is a multilingual extension of the original HumanEval benchmark, designed to evaluate the capabilities of code generation models across multiple programming languages. It consists of 164 handcrafted programming problems, each translated into five major languages: C++, Java, JavaScript, Go, and Python. This results in a total of 820 problem-solution pairs, supporting both code generation and code translation tasks. HumanEval-X allows for the assessment of a model's ability to generate functionally correct code and to translate code between different languages, using test cases to verify the correctness of the generated code. This benchmark facilitates a more comprehensive understanding of pre-trained multilingual code generation models and their potential applications in real-world programming tasks.

To assess the coding prowess of InternLM2, we conduct a series of experiments utilizing the widely recognized benchmarks MBPP~\citep{austin2021program} and HumanEval~\citep{chen2021evaluating}. Furthermore, to gauge the aptitude of code generation models for multiple programming languages, we extend our evaluation to include MBPP-CN, a Chinese adaptation of MBPP, and HumanEval-X, a multilingual expansion of the original HumanEval benchmark. 

As depicted in Figure~\ref{tab:chat_code_results}, the InternLM2 model series achieve leading performance, especially on HumanEval, MBPP, and MBPP-CN, where the InternLM2-Chat-20B model surpasses the previous state-of-the-art by more than 10\%, underscoring the InternLM2 series' exceptional proficiency in code generation tasks. Additionally, the InternLM2-Chat-20B model exhibits substantial improvement over the InternLM2-Chat-7B model in MBPP-CN benchmarks, yet it shows a slight decline in performance on HumanEval-X. This phenomenon might stem from the InternLM2-Chat-20B model being finely tuned for Chinese at the expense of its effectiveness in other languages.

\begin{table*}[!thb]
\centering

\resizebox{0.9\linewidth}{!}{
\tablestyle{10pt}{1.2}
\begin{tabular}{l|cccc}
\toprule
\textbf{Models} & \textbf{HumanEval} & \textbf{HumanEval-X} & \textbf{MBPP}  & \textbf{MBPP-CN}   \\
& \textit{4-shot} & \textit{5-shot} & \textit{5-shot} & \textit{0-shot} \\
\multicolumn{5}{c}{\greencolor{\textit{  $\leq 7$B Models}}} \\
Llama2-7B   &  14.6  &  11.2  &  28.8  &  16.0  \\
Mistral-7B-v0.1   &  27.4  &  28.5  &  47.5  &  35.0  \\
Baichuan2-7B-Base   &  22.0  &  16.1  &  35.0  &  23.2  \\
Qwen-7B-Chat &  36.0  &  24.4  &  33.9  &  27.0 \\
ChatGLM3-6B-Base   &  45.1  &  38.3  &  \textbf{\underline{68.9}}  &  \underline{50.0}  \\
\midrule
InternLM2-7B   &  \underline{56.1}  &  \underline{39.6}  &  51.8  &  45.4  \\
InternLM2-7B-Base   &  31.1  &  28.8  &  54.1  &  40.6  \\
\multicolumn{5}{c}{\bluecolor{\textit{ $13\sim 20$B Models}}} \\
Llama2-13B   &  18.9  &  17.2  &  38.9  &  25.2  \\
Mixtral-8x7B-v0.1   &  32.3  &  38.3  &  59.5  &  43.8  \\
Baichuan2-13B-Base   &  23.2  &  19.5  &  44.0  &  30.6  \\
Qwen-14B   &  30.5  &  31.0  &  52.9  &  41.0  \\
\midrule
InternLM2-20B   &  48.8  &  \textbf{\underline{48.2}}  &  \underline{63.0}  &  \textbf{\underline{53.2}}  \\
InternLM2-20B-Base   &  32.3  &  31.5  &  59.9  &  44.4  \\
\bottomrule
\end{tabular}}
\caption{\textbf{Comparison of Base Models on Python Coding}. The model name in \textbf{bold} indicates the top performer, while an \underline{underline} signifies the leading model within a similar parameter size group.}
\label{tab:chat_code_results}
\end{table*}

\begin{table*}[!thb]
\centering

\resizebox{0.9\linewidth}{!}{
\tablestyle{10pt}{1.2}
\begin{tabular}{l|cccc}
\toprule
\textbf{Models} & \textbf{HumanEval} & \textbf{HumanEval-X} & \textbf{MBPP}  & \textbf{MBPP-CN}   \\
& \textit{4-shot} & \textit{5-shot} & \textit{5-shot} & \textit{0-shot} \\
\multicolumn{5}{c}{\greencolor{\textit{  $\leq 7$B Models}}} \\
Llama2-7B-Chat &  15.2  &  10.6  &  30.4  &  17.8 \\
Mistral-7B-Instruct-v0.2  &  35.4  &  27.1  &  23.7  &  17.6 \\
Baichuan2-7B-Chat &  17.7  &  15.4  &  37.7  &  20.8  \\
Qwen-7B-Chat &  36.0  &  24.4  &  33.9  &  27.0 \\
ChatGLM3-6B &  53.1  &  17.6  &  \underline{54.9}  &  38.2 \\
\midrule
InternLM2-Chat-7B-SFT &  \underline{61.6}  &  \underline{\textbf{43.9}}  &  47.5  &  44.4  \\
InternLM2-Chat-7B &  59.2  &  41.7  &  52.5  &  \underline{46.4}   \\
\multicolumn{5}{c}{\bluecolor{\textit{ $13\sim 20$B Models}}} \\
Llama2-13B-Chat &  8.5  &  12.9  &  19.8  &  22.2  \\
Mixtral-8x7B-Instruct-v0.1 &  32.3  &  38.3  &  59.5  &  43.8 \\
Baichuan2-13B-Chat &  19.5  &  18.3  &  40.9  &  27.8 \\
Qwen-14B-Chat &  41.5  &  29.9  &  29.2  &  27.6   \\
\midrule
InternLM2-Chat-20B-SFT &  67.1  &  \underline{42.0}  &  63.4  &  52.2   \\
InternLM2-Chat-20B &  \underline{\textbf{67.7}}  &  39.8  &  \underline{\textbf{65.8}}  &  \underline{\textbf{54.0}}  \\
\bottomrule
\end{tabular}}
\caption{\textbf{Comparison of Chat Models on Python Coding}. The model name in \textbf{bold} indicates the top performer, while an \underline{underline} signifies the leading model within a similar parameter size group.}
\label{tab:chat_code_results}
\end{table*}

\subsubsection{Performance Before and After Enhancement Training}
\begin{figure*}[!htbp]
  \centering
  \includegraphics[scale=0.5]{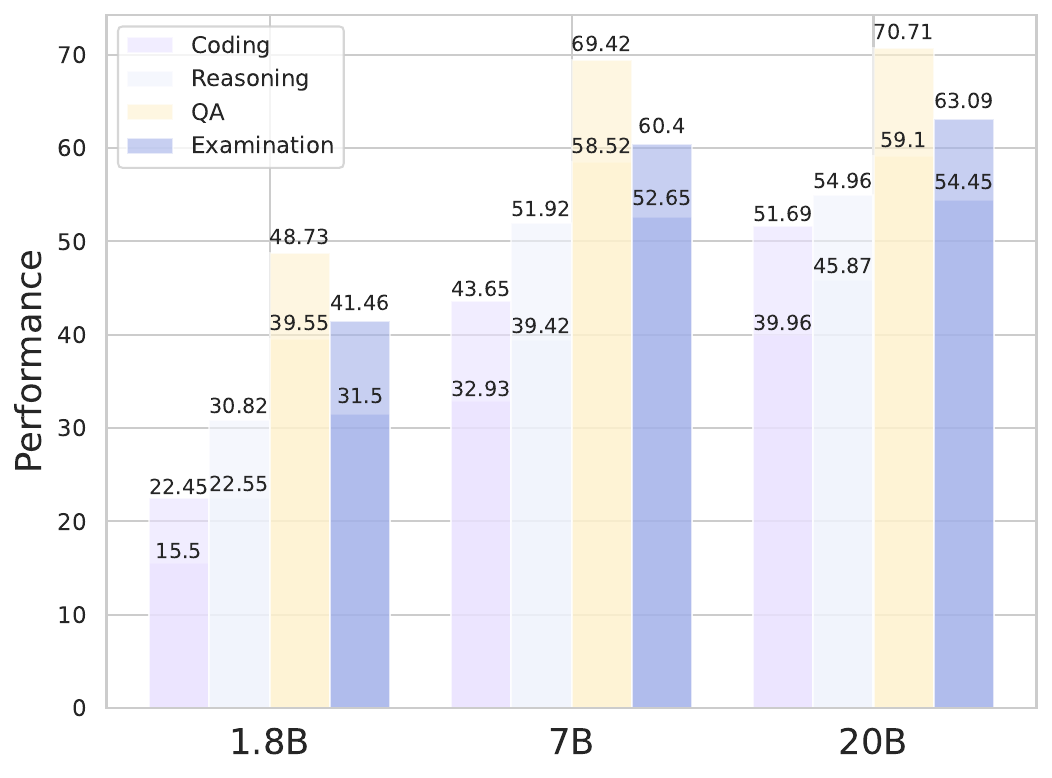}
  \caption{The performance before and after the capability specific enhancement training phase, the darker areas mean the performance before the training.}
  \label{fig:before_and_after_enhancement}
\end{figure*}

Through results in previous parts, we can clearly see that InternLM2 with capability-specific enhancement training consistently outperforms its counterpart. The overall comparison of model performance before and after this phase is presented in Figure~\ref{fig:before_and_after_enhancement}, where the scores for various capabilities are derived from the average of multiple evaluation sets.

\textbf{Coding:} HumanEval and MBPP.

\textbf{Reasoning:} MATH, GSM8k, SummEdits~\citep{DBLP:conf/emnlp/LabanKAFXJW23} and BBH.

\textbf{QA:} HellaSwag, PIQA~\citep{DBLP:conf/aaai/BiskZLGC20}, WinoGrande, OpenBookQA~\citep{DBLP:conf/emnlp/MihaylovCKS18}, NaturalQuestions and TriviaQA.

\textbf{Examination:} MMLU, AGIEval and C-Eval.

It is evident that there has been a significant improvement in these capabilities. Furthermore, we also compared the performance of the base models before and after undergoing enhancement training when subjected to Supervised Fine-Tuning~(SFT) in Table~\ref{table:ablation_boost_sft}. The capability dimensions are aligned with the categorization adopted in OpenCompass~\footnote{\url{https://rank.opencompass.org.cn/leaderboard-llm-v2}}. It can be observed that SFT models trained with capability-specific enhancement achieve better performance across various capability dimensions. The SFT performance in other parts of this report is based on the base model before the enhancement training.

\begin{table}[h!]
\centering
\begin{tabularx}{\textwidth}{
    >{\centering\arraybackslash\hsize=1.4\hsize}X
    >{\centering\arraybackslash\hsize=0.4\hsize}X
    >{\centering\arraybackslash\hsize=0.4\hsize}X
    >{\centering\arraybackslash\hsize=0.4\hsize}X
    >{\centering\arraybackslash\hsize=0.4\hsize}X
    >{\centering\arraybackslash\hsize=0.4\hsize}X
    >{\centering\arraybackslash\hsize=0.4\hsize}X}
    
    \toprule
Model                         & Language & Knowledge & Reason & Code  \\
\midrule
SFT from InternLM2-7B-Base & 66.64 & 58.35 & 69.30 & 38.79 \\
%\hline
SFT from InternLM2-7B  & \textbf{69.28} & \textbf{62.11} &  \textbf{75.18}  & \textbf{41.40} \\
\bottomrule
\end{tabularx}
\caption{Comparative performance of a 7B SFT model trained from InternLM2-7B-Base and from InternLM2-7B. }
\label{table:ablation_boost_sft}
\end{table}

\subsubsection{Long-context Modeling}
\input{sections/evaluation/long_context_subsub}

\input{sections/evaluation/agent}

\subsection{Performance on Alignment}
Although LLMs' objective capabilities improve through pretraining and SFT, their answer styles may not align with human preferences, necessitating further enhancement with RLHF to improve alignment. Therefore, assessing the alignment capabilities is crucial to determine whether LLMs truly meet human needs.

In this section, we evaluate InternLM2's performance on several popular subjective alignment datasets, comparing the performance of SFT and RLHF models. As we can see in Table \ref{tab:alignment_model_results}, InternLM2's overall performance in alignment tasks achieved SOTA or near-SOTA results on multiple benchmarks, demonstrating a high level of consistency between the subjective outputs of InternLM2 series models and human preferences. We detail the model performance on each dataset separately below. Additionally, the prompt templates used for subjective alignment tasks on each dataset are presented in the Appendix.

\begin{table*}[!hthb]
\centering

\resizebox{\linewidth}{!}{
\begin{tabular}{lccccc}
\hline
\diagbox{Models}{Datasets } & \textbf{AlpacaEval} & \textbf{MTBench} & \textbf{CompassArena} & \textbf{AlignBench} & \textbf{IFEval} \\
& \textit{WinRate} & \textit{0-10 Score} & \textit{WinRate} & \textit{0-10 Score} & \textit{0-100 Acc}\\
\hline
\multicolumn{6}{c}{\purplecolor{\textit{\ding{72}API Models}}} \\
GPT-3.5 & 9.6 & 8.4 & 24.7 & 5.7 & - \\
\multicolumn{6}{c}{\greencolor{\textit{$\triangle$ $\sim7B Models$}}} \\
ChatGLM3-6B & - & - & 17.2 & 5.0 & 33.7 \\
Llama2-7B-Chat & 5 & 6.3 & 6.5 & - & 44.6 \\
Baichuan2-7B-Chat & - & - & 16.1 & 5.1 & 42.1 \\
Mistral-7B-Instruct-v0.2 & \underline{14.7} & - & 14.2 & - & \underline{\textbf{57.9}} \\
Qwen-7B-Chat & - & - & 15.8 & 4.7 & 37.3 \\
\midrule
InternLM2-Chat-7B-SFT & 6.9 & 7.1 & 14.1 & 4.9 & 48.5  \\
InternLM2-Chat-7B & 11.3 & \underline{7.7} & \underline{28.7} & \underline{6.1} & 45.9 \\
\multicolumn{6}{c}{\bluecolor{\textit{$\heartsuit$ $\sim20B Models$}}} \\
Llama2-13B-Chat & 7.7 & 6.7 & 9.8 & - & 46.1 \\
Baichuan2-13B-Chat & - & - & 20.5 & 5.3 & 42.5 \\
Mixtral-8x7B-Instruct-v0.1 & - & \underline{\textbf{8.3}} & 18.9 & - & \underline{56.5} \\
Qwen-14B-Chat & 7.5 & 7.0 & 24.8 & 5.4 & 43.8 \\
\midrule
InternLM2-Chat-20B-SFT & 8.0 & 7.3 & 15.2 & 5.3 & 48.7   \\
InternLM2-Chat-20B & \underline{\textbf{21.8}} & 7.9 & \underline{\textbf{31.4}} & \underline{\textbf{6.8}} & 42.2  \\
\hline
\end{tabular}}
\caption{\textbf{Comparison of Models on Alignment Benchmarks}. Models are categorized by their parameter size and type, highlighting top performers in each category with \textbf{bold} for overall leaders and \underline{underline} for leaders within their parameter group.}
\vspace{-2mm}
\label{tab:alignment_model_results}
\end{table*}

\subsubsection{English Subjective Evaluation}
\paragraph{AlpacaEval}

AlpacaEval\citep{alpaca_eval} is a single-turn question-answer dataset with 805 questions. Its main purpose is to assess the helpfulness of model responses to humans, reflecting the alignment with human intentions through adversarial evaluation. AlpacaEval(v2) establishes a baseline model and uses GPT4-Turbo as the judge model to compare the baseline's answers with those of the model under evaluation. It selects the model that better aligns with human preferences and calculates the win rate.
Instead of directly collecting the judge model's responses, AlpacaEval(v2) employs logit probabilities to analyze the judge model's preferences statistically. These preferences are then used as a weighting factor in the calculation of the weighted win rate.
As shown in Table~\ref{tab:alignment_model_results}, InternLM2-20B achieved a win rate of 21.8, marking the highest SOTA result among the compared models and demonstrating its superior alignment performance. Moreover, the table indicates that the RLHF model outperforms the SFT model in terms of win rate, which underscores the effectiveness of the alignment strategy.
\paragraph{MTBench}
MTBench\citep{zheng2023judging} is a two-round conversation dataset consisting of 80 questions that span eight dimensions: reasoning, roleplay, math, coding, writing, humanities, STEM, and information extraction. Each dimension contains 10 questions, which are subjected to two rounds of questioning. Initially, the model's ability to answer basic questions is tested; subsequently, it must follow additional, specific instructions to refine its previous response. Scores are assigned on a 1-10 scale by a judge model.
As we can see in Table~\ref{tab:alignment_model_results}, InternLM2 achieves leading scores in both the 7B and 20B versions, with 7.7 and 7.9 respectively, demonstrating its reliable multi-turn conversational ability.

\begin{figure*}[!htbp]
    \centering
    \includegraphics[scale=0.4]{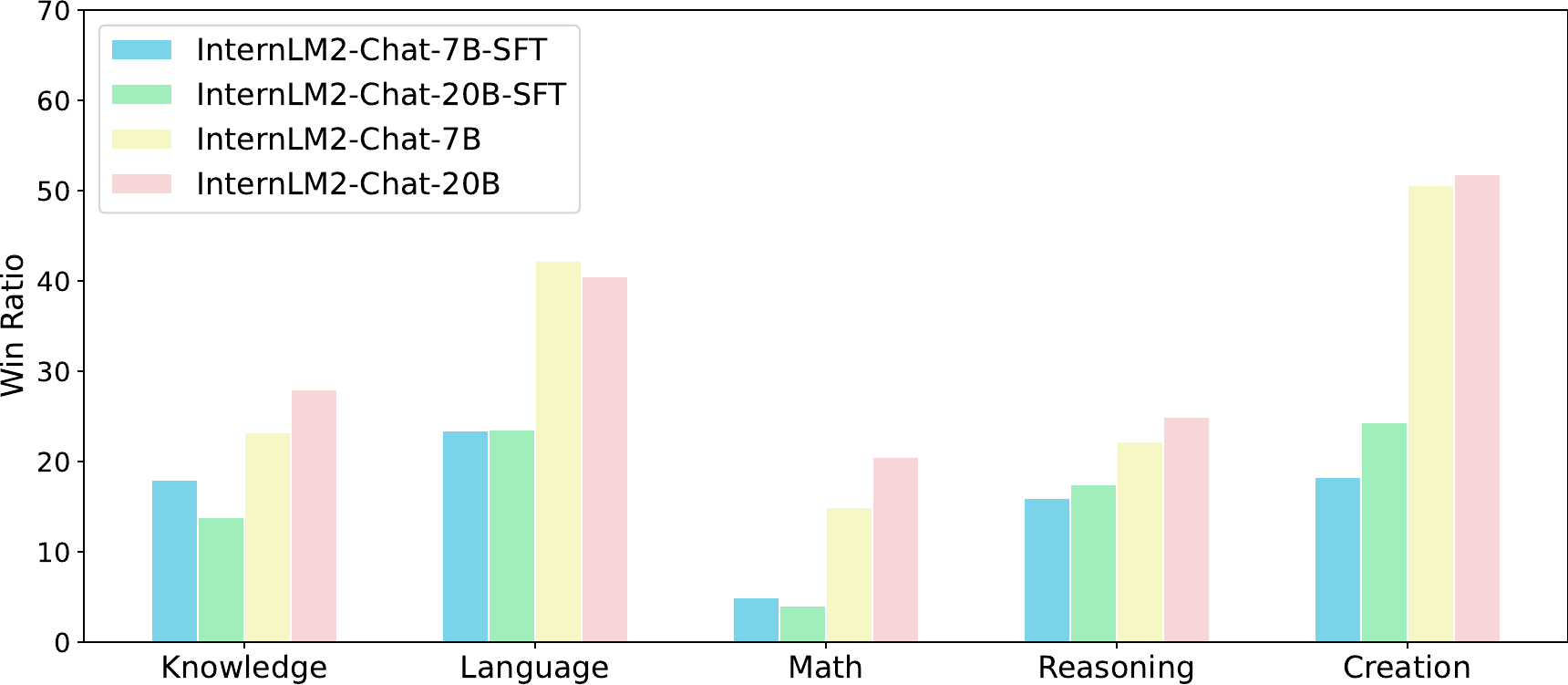}
    \caption{Detail Results on CompassArena of InternLM2 Series.}
    \label{fig:compass_arena}
\end{figure*}
\subsubsection{Chinese Subjective Evaluation}
\paragraph{CompassArena}
CompassArena comprises 520 Chinese questions, encompassing knowledge, language, mathematics, reasoning, and creativity. Like AlpacaEval, it calculates the win rate of two models judged by GPT4-Turbo and employs double-blind testing by swapping the model order to mitigate position bias. As indicated in Table~\ref{tab:alignment_model_results}, InternLM2 secures the highest win rate in the 7B version (28.7) and the 20B version (31.4). 
Note that the performance gap between InternLM2's 7B and 20B versions is relatively small. However, when compared to SFT models, InternLM2's RLHF model shows a significant improvement in performance. This suggests that the RLHF strategy substantially enhances InternLM2's alignment with human preferences, rather than just increasing the model size. Furthermore, the results broken down by category in Figure~\ref{fig:compass_arena} reveal that the InternLM2 series possesses exceptionally strong Chinese creativity and language abilities, with a win rate that rivals that of GPT4-Turbo.

\paragraph{AlignBench}
AlignBench\citep{liu2023alignbench} is a Chinese subjective dataset comprising 683 question-answer pairs, categorized into eight areas, including Fundamental Language Ability, Advanced Chinese Understanding, Task-oriented Role Play, among others, covering various scenarios such as physics, history, music, and law. The dataset utilizes an in-house Judge Model called CritiqueLLM\citep{ke2023critiquellm} for evaluation. CritiqueLLM provides scores from 1 to 10 across various dimensions for each question and issues a final score. We present these final scores, as provided by CritiqueLLM, in Table~\ref{tab:alignment_model_results} and detailed scores across different categories in Table~\ref{tab:detail_alignbench}. As shown in Table~\ref{tab:alignment_model_results}, InternLM2 achieves SOTA scores in both the 7B and 20B versions, with 6.1 and 6.8 respectively, outperforming GPT-3.5's score of 5.7.
Moreover, both the 7B and 20B versions of InternLM2 exhibit significant performance enhancements post-RLHF compared to SFT models, underscoring the effectiveness of our RLHF strategy in improving the model's alignment with human preferences. An analysis of the detailed scores reveals areas for improvement in mathematical and reasoning abilities; however, InternLM2 excels in question-answering and role-playing tasks, offering strong subjective performance.

\begin{table*}[!hthb]
\centering
\resizebox{0.8\linewidth}{!}{
\begin{tabular}{lccccccccc}
\hline
Models& \textbf{K
} & \textbf{U
} & \textbf{L
} & \textbf{M
} & \textbf{W
} & \textbf{QA
} & \textbf{RP
} & \textbf{RE
}  \\
\hline
InternLM2-Chat-7B
 & 7.12                                          & 7.26                                                  & 6.34                                                & 4.55                               & 7.73                                   & 8.00                                        & 7.83                                           & 5.22                                     \\
InternLM2-Chat-7B-SFT
 & 5.83                                          & 5.81                                                  & 5.63                                                & 3.82                               & 6.25                                   & 6.89                                        & 6.46                                           & 3.51                                     \\
 InternLM2-Chat-20B& 8.02                                          & 7.88                                                  & 7.65                                                & 5.38                               & 7.93                                   & 8.26                                        & 8.09                                           & 5.76    \\
InternLM2-Chat-20B-SFT
 & 6.41                                          & 6.10                                                  & 5.34                                                & 4.28                               & 6.69                                   & 6.47                                        & 6.66                                           & 4.35       \\

\hline
\end{tabular}}
\caption{\textbf{Detail Scores of InternLM2 Series on AlignBench.} K: Knowledge, U: Understadning, L: Language, M: Mathematics, W: Writing, RP: Role Play, RE: Reasoning.}
\vspace{-2mm}
\label{tab:detail_alignbench}
\end{table*}

\subsubsection{Instruct Following Evaluation}
\paragraph{IFEval}

IFEval\citep{zhou2023instruction} is designed to test models' instruction following ability, requiring responses to adhere to specific patterns, such as letter case restrictions and keyword limitations. IFEval encompasses 25 distinct types of instructions, constructing 541 questions that follow instructions, and employs rule-based evaluation to assess the correctness of models' responses for each question. Utilizing various statistical methods, IFEval offers four types of accuracy scores: prompt-level strict, prompt-level loose, instance-level strict, and instance-level loose. We present the average results of these four scores in Table~\ref{tab:alignment_model_results}.
Although English-based models like Llama2 and Mistral generally outperform Chinese-based models on IFEval, InternLM2 ranks second and third in the 7B phase (48.5) and the 13-20B phase (48.7), respectively. This indicates that, despite the challenges of instruction following tasks, InternLM2 maintains a leading position among models of similar size.

\subsubsection{Ablation Study of Conditional Reward Model} 

To verify the impact of conditional system prompts, we compare the performance of the reward model trained on a heterogeneous mix of data from different domains, with and without using conditional system prompts. As illustrated in Table~\ref{table:ablation_condition}, the absence of system prompts results in a significant decrease in precision across several public datasets, including scenarios such as helpful and harmless conversations~\citep{bai2022training}, content summaries~\citep{stienon2020learning}, math problems~\citep{lightman2023lets}, and Reddit replies~\citep{pmlr-v162-ethayarajh22a}. Conversely, including system prompts leads to markedly higher precision in these areas.

\begin{table}[h!]
\centering
\begin{tabularx}{\textwidth}{
    >{\centering\arraybackslash\hsize=1\hsize}X|
    >{\centering\arraybackslash\hsize=0.85\hsize}X
    >{\centering\arraybackslash\hsize=0.95\hsize}X
    >{\centering\arraybackslash\hsize=0.9\hsize}X
    >{\centering\arraybackslash\hsize=0.4\hsize}X
    >{\centering\arraybackslash\hsize=0.4\hsize}X
    >{\centering\arraybackslash\hsize=0.4\hsize}X}
    \toprule
  & Anthropic Helpful-base & Anthropic Helpful-online & Anthropic Harmless-base & OpenAI Summ. & Stanford SHP & PRM 800k \\
\midrule

UltraRM  & 75.02 & 65.34 &  37.02  & 74.48 & 74.36 &  48.77 \\

QwenRM  & 73.98 & 64.57 &  -  & 69.99 & 60.10 &  70.52 \\

Ours w/o cond. & 73.26 & 64.45 & 72.48 &68.25 & 69.24 & 72.11 \\
%\hline
Ours w/ cond. & 75.10 & 66.98 & 73.19 & 75.37 & 75.76 & 77.18 \\
\bottomrule
\end{tabularx}
\caption{Comparative performance of UltraRM-13B\citep{cui2023ultrafeedback}, QwenRM\citep{DBLP:journals/corr/abs-2309-16609}, and ours 7B reward model trained with and without conditional system prompts. The table demonstrates a marked improvement in precision when conditional prompts are used.}
\label{table:ablation_condition}
\end{table}

\subsection{Discussion on Data Contamination}
\label{sec:contamination}

\input{sections/evaluation/contamination_table}

We evaluate the language modeling (LM) loss on samples (a sample is a concatenation of question and answer) from GSM8K dataset for several foundation models, results are shown in Table~\ref{tab:language_conta_results}. For each LLM, we compare LM loss on the training split (\(L_{train}\)), the test split (\(L_{test}\)), and a specially curated reference set (\(L_{ref}\)), generated by GPT-4, designed to mimic the GSM8K dataset. We also report two key metrics: \(\Delta_1 = L_{test} - L_{ref}\), serving as an indicator of potential test data leakage during the training of the LLM, i.e., a lower value suggests possible leakage; and \(\Delta_2 = L_{test} - L_{train}\), which measures the degree of overfitting on the training split of the dataset. A higher value of \(\Delta_2\) implies excessive overfitting. Outliers for both \(\Delta_1\) and \(\Delta_2\) are highlighted in gray.

\section{Conclusion}
In this report, we present the InternLM2 large language model, which demonstrates exceptional performance in both subjective and objective evaluations. InternLM2 has been trained on over 2T of high-quality pre-training corpora, covering model sizes of 1.8B, 7B, and 20B, making it suitable for a variety of scenarios. To better support long contexts, InternLM2 employs GQA to reduce inference costs, and has been additionally trained on up to 32k contexts. Besides open-sourcing the model itself, we also make available the checkpoints from various phases of the training process, facilitating studies by future researches.

In addition to open-sourcing the model, we provide a detailed description of how we trained InternLM2, including the training framework, pre-training text data, pre-training code data, pre-training long text data, and alignment data. Furthermore, to address preference conflicts encountered during the RLHF process, we propose Conditional Online RLHF to harmonize various preferences. This information can offer insights into how to prepare pre-training data and how to train larger models more effectively.

\bibliography{main}
\bibliographystyle{colm2024_conference}

\newpage
\appendix
\input{sections/appendix}

\end{document}

%% file: math_commands.tex
\usepackage{amsmath,amsfonts,bm}

\def\eqref#1{equation~\ref{#1}}
\def\1{\bm{1}}

\DeclareMathAlphabet{\mathsfit}{\encodingdefault}{\sfdefault}{m}{sl}
\SetMathAlphabet{\mathsfit}{bold}{\encodingdefault}{\sfdefault}{bx}{n}

\newcommand{\sigmoid}{\sigma}

%% file: sections/evaluation/examination_table.tex
\newcommand\redcolor[1]{\cellcolor{gray!40!red!30}{#1}}
\newcommand\bluecolor[1]{\cellcolor[HTML]{EAE2FE}{#1}}
\newcommand\greencolor[1]{\cellcolor[HTML]{DFF5E5}{#1}}
\newcommand\yellowcolor[1]{\cellcolor[HTML]{FEF1CE}{#1}}
\newcommand\purplecolor[1]{\cellcolor[HTML]{A8B3E3}{#1}}

% 基座模型

\begin{table*}[!thb]
\centering

\resizebox{\linewidth}{!}{
\tablestyle{10pt}{1.2}
\begin{tabular}{l|ccccc}
\toprule
\textbf{Models} & \textbf{MMLU} & \textbf{CMMLU} & \textbf{C-Eval} & \textbf{AGIEval} & \textbf{GAOKAO} \\
& \textit{5-shot} & \textit{5-shot} & \textit{5-shot} & \textit{0-shot} & \textit{0-shot}\\
\hline
\multicolumn{6}{c}{\greencolor{\textit{  $\leq 7$B Models}}} \\
Llama2-7B & 46.8 & 31.9 & 32.4 & 21.3 & 19.0 \\
% Gemma-7B & \\
Mistral-7B-v0.1 & 64.0 & 44.6 & 47.5 & 32.9 & 28.7 \\
Baichuan2-7B-Base & 54.7 & 57.0 & 56.2 & 34.5 & 34.7 \\
ChatGLM3-6B-Base & 62.7 & \underline{66.5} & \underline{67.2} & 47.5 & \underline{59.4} \\
Qwen-7B & 59.7 & 62.5 & 63.1 & 45.6 & 52.8 \\
% Qwen1.5-7B & 62.2 & \underline{71.8} & \underline{73.5} & \underline{50.8} & \underline{70.5} \\
\midrule
InternLM2-7B-Base & 64.0 & 63.0 & 60.7 & 38.7 & 40.7 \\
InternLM2-7B & \underline{65.8} & 66.3 & 65.8 & \underline{49.9} & 58.6 \\
\hline
\multicolumn{6}{c}{\bluecolor{\textit{ $13\sim 20$B Models}}} \\
Llama2-13B & 55.8 & 38.8 & 40.1 & 27.9 & 18.4 \\
Mixtral-8x7B-v0.1 & \textbf{\underline{71.8}} & 53.3 & 55.4 & 40.9 & 32.3 \\
Baichuan2-13B-Base & 59.6 & 61.3 & 59.2 & 37.4 & 45.6 \\
Qwen-14B & 67.9 & \textbf{\underline{70.1}} & \textbf{\underline{71.8}} & 52.0 & \textbf{\underline{62.5}} \\
% Qwen1.5-14B & 69.2 & \textbf{\underline{76.9}} & \textbf{\underline{77.9}} & \textbf{\underline{56.4}} & \textbf{\underline{75.5}} \\
\midrule
InternLM2-20B-Base & 64.0 & 63.0 & 60.7 & 38.7 & 40.7 \\
InternLM2-20B & 67.7 & 68.7 & 68.5 & \textbf{\underline{53.0}} & 57.1 \\
\bottomrule
\end{tabular}}
\caption{\textbf{Comparison of Base Models on Comprehensive Examination}. The model name in \textbf{bold} indicates the top performer, while an \underline{underline} signifies the leading model within a similar parameter size group.}
% \vspace{-2mm}
\label{tab:exam_base_results}
\end{table*}

% 对话模型
\begin{table*}[!thb]
\centering
\centering

\resizebox{\linewidth}{!}{
\tablestyle{10pt}{1.2}
\begin{tabular}{l|ccccc}
\toprule
\textbf{Models} & \textbf{MMLU} & \textbf{CMMLU} & \textbf{C-Eval} & \textbf{AGIEval} & \textbf{GAOKAO} \\
& \textit{5-shot} & \textit{5-shot} & \textit{5-shot} & \textit{0-shot} & \textit{0-shot}\\
\hline
\multicolumn{6}{c}{\purplecolor{\textit{ API Models}}} \\
GPT-3.5 & 69.1 & 53.9 & 52.5 & 39.9 & 51.1 \\
\multicolumn{6}{c}{\greencolor{\textit{  $\leq 7$B Models}}} \\
Llama2-7B-Chat & 48.2 & 30.7 & 34.9 & 28.5 & 19.8 \\
% Gemma-7B-IT &  &  &  &  &  \\
Mistral-7B-Instruct-v0.2 & 59.2 & 42.0 & 42.4 & 34.5 & 28.6 \\
Baichuan2-7B-Chat & 50.1 & 53.4 & 53.9 & 35.3 & 37.5 \\
ChatGLM3-6B & 58.0 & 57.8 & 59.1 & 44.2 & 56.3 \\
Qwen-7B-Chat & 57.1 & 57.9 & 59.8 & 39.7 & \underline{62.1} \\
% Qwen1.5-7B-Chat & 61.0 & \underline{68.0} & \underline{70.6} & 44.9 & \underline{69.9} \\
\midrule
InternLM2-Chat-7B-SFT & \underline{63.8} & \underline{63.2} & \underline{60.9} & \underline{49.0} & 57.3 \\
InternLM2-Chat-7B & 63.7 & 63.0 & 60.8 & 47.2 & 58.0 \\
\multicolumn{6}{c}{\bluecolor{\textit{ $13\sim 20$B Models}}} \\
Llama2-13B-Chat & 54.1 & 33.8 & 35.0 & 30.9 & 23.2 \\
Mixtral-8x7B-Instruct-v0.1 & \underline{\textbf{70.3}} & 50.6 & 54.0 & 41.7 & 42.6 \\
Baichuan2-13B-Chat & 56.6 & 54.8 & 56.3 & 40.0 & 45.8 \\
Qwen-14B-Chat & 66.7 & \underline{\textbf{68.1}} & \underline{\textbf{71.5}} & 46.5 & \underline{\textbf{76.3}} \\
% Qwen1.5-14B-Chat & 67.2 & \underline{\textbf{75.1}} & \underline{\textbf{76.0}} & 50.6 & \underline{\textbf{80.3}} \\
\midrule
InternLM2-Chat-20B-SFT & 66.5 & 65.3 & 63.7 & \underline{\textbf{51.2}} & 58.6 \\
InternLM2-Chat-20B & 66.5 & 65.1 & 63.0 & 50.3 & 58.3 \\
\bottomrule
\end{tabular}}
\caption{\textbf{Comparison of Chat Models on Comprehensive Examination}. The model name in \textbf{bold} indicates the top performer, while an \underline{underline} signifies the leading model within a similar parameter size group.}
% \vspace{-2mm}
\label{tab:exam_chat_results}
\end{table*}

%% file: sections/evaluation/language_table_1.tex
% \newcommand\redcolor[1]{\cellcolor{gray!40!red!30}{#1}}
% \newcommand\bluecolor[1]{\cellcolor{gray!40!blue!30}{#1}}
% \newcommand\greencolor[1]{\cellcolor{gray!40!green!30}{#1}}
% \newcommand\yellowcolor[1]{\cellcolor{gray!20!yellow!40}{#1}}
% \newcommand\purplecolor[1]{\cellcolor{gray!20!purple!30}{#1}}

% 基座模型
\begin{table*}[!thb]
\centering

\resizebox{\linewidth}{!}{
\tablestyle{10pt}{1.2}
\begin{tabular}{l|ccccc}
\toprule
\textbf{Models} & \textbf{TriviaQA} & \textbf{NaturalQuestions} & \textbf{C3} & \textbf{RACE-High} & \textbf{FLORES} \\
& \textit{4-shot} & \textit{5-shot} & \textit{5-shot} & \textit{0-shot} & \textit{0-shot}\\
\hline
\multicolumn{6}{c}{\greencolor{\textit{  $\leq 7$B Models}}} \\
Llama2-7B & 55.9 & 28.8 & 43.8 & 37.6 & 6.0 \\
% Gemma-7B &  &  &  &  &  \\
Mistral-7B-v0.1 & \underline{68.9} & 31.1 & 54.6 & 69.1 & \underline{6.9} \\
Baichuan2-7B-Base & 51.0 & 24.0 & 64.6 & 63.2 & 5.8 \\
Qwen-7B & 54.2 & 22.7 & 71.4 & 83.9 & 3.3 \\
% Qwen1.5-7B & 57.5 & 22.2 & 74.7 & \underline{85.2} & 5.7 \\
ChatGLM3-6B-Base & 63.9 & 38.7 & \underline{78.9} & \underline{84.8} & 0.9 \\
\midrule
InternLM2-7B-Base & 57.0 & 28.4 & 61.9 & 74.2 & 5.5 \\
InternLM2-7B & 58.9 & \underline{42.2} & 74.1 & 81.4 & 5.9 \\
\hline
\multicolumn{6}{c}{\bluecolor{\textit{ $13\sim 20$B Models}}} \\
Llama2-13B & 60.7 & 34.5 & 47.5 & 59.1 & 7.2 \\
Mixtral-8x7B-v0.1 & \underline{\textbf{77.6}} & 39.6 & 59.1 & 80.2 & \underline{\textbf{8.8}} \\
Baichuan2-13B-Base & 54.7 & 27.5 & 65.6 & 75.6 & 6.4 \\
Qwen-14B & 62.5 & 37.1 & \underline{\textbf{80.6}} & \underline{\textbf{90.3}} & 5.9 \\
% Qwen1.5-14B & 59.3 & 22.4 & 77.7 & 89.1 & 6.5 \\
\midrule
InternLM2-20B-Base & 63.7 & 30.3 & 67.6 & 75.6 & 6.0 \\
InternLM2-20B & 60.1 & \underline{\textbf{44.6}} & 79.1 & 72.9 & 6.5 \\
\bottomrule
\end{tabular}}
\caption{\textbf{Comparison of Base Models on Language \& Knowledge}. The model name in \textbf{bold} indicates the top performer, while an \underline{underline} signifies the leading model within a similar parameter size group.}
% \vspace{-2mm}
\label{tab:language_base_results}
\end{table*}

% 对话模型

\begin{table*}[!thb]
\centering

\resizebox{\linewidth}{!}{
\tablestyle{10pt}{1.2}
\begin{tabular}{l|ccccc}
\toprule
\textbf{Models} & \textbf{TriviaQA} & \textbf{NaturalQuestions} & \textbf{C3} & \textbf{RACE-High} & \textbf{FLORES} \\
& \textit{4-shot} & \textit{5-shot} & \textit{5-shot} & \textit{0-shot} & \textit{0-shot}\\
\hline
% \multicolumn{7}{c}{\purplecolor{\textit{\ding{72} $> 20$B}}} \\
\multicolumn{6}{c}{\purplecolor{\textit{ API Models}}} \\
GPT-3.5 & 69.1 & 53.9 & 52.5 & 39.9 & 51.1 \\
% \hline
\multicolumn{6}{c}{\greencolor{\textit{  $\leq 7$B Models}}} \\
Llama2-7B-Chat & 46.8 & 19.6 & 51.7 & 3.0 & 5.3 \\
% Gemma-7B-IT &  &  &  &  &  \\
Mistral-7B-Instruct-v0.2 & 35.0 & 8.1 & 66.9 & 71.2 & 12.8 \\
Baichuan2-7B-Chat & 37.6 & 12.8 & 78.5 & 67.5 & 11.0 \\
Qwen-7B-Chat & 46.1 & 18.6 & 84.4 & 74.8 & 12.9 \\
% Qwen1.5-7B-Chat & 44.6 & 12.7 & 88.6 & 82.3 & 12.2 \\
ChatGLM3-6B & 38.1 & 14.0 & 79.3 & 79.1 & 9.7 \\
\midrule
InternLM2-Chat-7B-SFT & \underline{51.4} & \underline{24.5} & \underline{91.7} & \underline{85.2} & \underline{15.5} \\
InternLM2-Chat-7B & 50.8 & 24.1 & 91.5 & 85.1 & 15.2 \\
% \hline
\multicolumn{6}{c}{\bluecolor{\textit{ $13\sim 20$B Models}}} \\
Llama2-13B-Chat & 54.4 & 25.2 & 56.9 & 19.2 & 6.6 \\
Mixtral-8x7B-Instruct-v0.1 & \underline{\textbf{57.7}} & 22.5 & 82.1 & 81.4 & 17.2 \\
Baichuan2-13B-Chat & 40.3 & 12.7 & 84.4 & 73.2 & 12.4 \\
Qwen-14B-Chat & 54.5 & 22.9 & 91.5 & 84.7 & \textbf{\underline{17.3}} \\
% Qwen1.5-14B-Chat & 54.9 & 15.3 & 93.3 & 86.4 & 15.4 \\
\midrule
InternLM2-Chat-20B-SFT & 55.5 & \textbf{\underline{27.8}} & 93.5 & \textbf{\underline{87.3}} & 17.0 \\
InternLM2-Chat-20B & 53.9 & 25.9 & \textbf{\underline{93.5}} & 87.1 & 16.9 \\
\bottomrule
\end{tabular}}
\caption{\textbf{Comparison of Chat Models on Language \& Knowledge}. The model name in \textbf{bold} indicates the top performer, while an \underline{underline} signifies the leading model within a similar parameter size group.}
% \vspace{-2mm}
\label{tab:language_chat_results}
\end{table*}

%% file: sections/evaluation/math_table_1.tex
\begin{table*}[!thb]
\centering

\resizebox{\linewidth}{!}{
% \tablestyle{10pt}{1.2}
\begin{tabular}{lccccc}
\hline
\textbf{Models} & \textbf{GSM8K} & \textbf{MATH} & \textbf{TheoremQA} & \textbf{MathBench-CN}& \textbf{MathBench-EN}\\
& \textit{4-shot} & \textit{4-shot}& \textit{0-shot} & \textit{0-shot\&4-shot}& \textit{0-shot\&4-shot}\\
\hline
\multicolumn{6}{c}{\greencolor{\textit{$\triangle$ $\sim7B Models$}}} \\
ChatGLM3-6B-Base & 60.7 & 19.2 & 6.5 & 32.5 & 29.0 \\
Baichuan2-7B-Base & 24.6 & 5.5 & 1.6 & 21.3 & 15.2 \\
Llama2-7B & 16.2 & 3.3 & 1.0 & 4.1 & 7.6 \\
Mistral-7B-v0.1 & 47.5 & 11.3 & 1.8 & 19.3 & 27.6 \\
Qwen-7B & 54.0 & 13.4 & 5.0 & 26.0 & 21.8 \\
% Qwen1.5-7B-Base & 55.7 & \underline{20.4} & 7.3 & \underline{43.5} & 29.8 \\
\midrule
InternLM2-7B-Base & 36.0 & 8.9 & 2.6 & 14.2 & 18.5 \\
InternLM2-7B & \underline{70.8} & 20.2 & \underline{10.5} & \underline{37.4} & \underline{33.9} \\
\multicolumn{6}{c}{\bluecolor{\textit{$\heartsuit$ $\sim20B Models$}}} \\
Llama2-13B & 28.7 & 4.9 & 2.6 & 10.6 & 18.8 \\
Baichuan2-13B-Base & 52.6 & 10.1 & 4.6 & 28.4 & 24.4 \\
Qwen-14B & 60.1 & 25.1 & 10.4 & \textbf{46.2} & 39.7 \\
% Qwen1.5-14B-Base & 64.8 & \textbf{29.3} & 10.4 & \textbf{54.9} & \textbf{48.2} \\
Mixtral-8x7B-v0.1 & 66.0 & 22.7 & 2.5 & 31.6 & \textbf{40.1} \\
\midrule
InternLM2-20B-Base & 54.3 & 13.6 & 3.3 & 27.0 & 26.8 \\
InternLM2-20B & \textbf{76.1} & 25.5 & \textbf{13.5} & 38.7 & 36.9 \\
\hline
\end{tabular}}
\caption{\textbf{Comparison of Base Models on Math Tasks.} The model name in \textbf{bold} indicates the top performer among Open-Source or API models, while an \underline{underline} signifies the leading model within a similar parameter size group.}
\vspace{-2mm}
\label{tab:base_math_result}
\end{table*}

%% file: sections/evaluation/math_table_2.tex
\begin{table*}[!thb]
\centering

\resizebox{\linewidth}{!}{
\begin{tabular}{lccccc}
\hline
\textbf{Models/Datasets} & \textbf{GSM8K} & \textbf{MATH} & \textbf{TheoremQA} & \textbf{MathBench-CN}& \textbf{MathBench-EN}\\
& \textit{4-shot} & \textit{4-shot}& \textit{0-shot} & \textit{0-shot\&8-shot}& \textit{0-shot\&8-shot}\\
\hline
\multicolumn{6}{c}{\purplecolor{\textit{\ding{72}API Models}}} \\
GPT-3.5 & 78.2 & 28.0 & 9.1 & 26.4 & 42.5 \\
\multicolumn{6}{c}{\greencolor{\textit{$\triangle$ $\sim7B Models$}}} \\
ChatGLM3-6B & 53.8 & 20.4 & 9.3 & 18.2 & 21.7 \\
Llama2-7B-Chat & 28.4 & 4.1 & 0.3 & 3.3 & 7.7 \\
Baichuan2-7B-Chat & 32.4 & 5.7 & 2.4 & 16.9 & 17.8 \\
Mistral-7B-Instruct-v0.2 & 48.3 & 8.6 & 2.3 & 14.5 & 28.3 \\
Qwen-7B-Chat & 44.1 & 12.0 & 5.8 & 31.2 & 29.0 \\
% Qwen1.5-7B-Chat & 56.9 & 9.3 & 5.4 & 38.0 & 34.2 \\
\midrule
InternLM2-Chat-7B-SFT & 68.8 & 23.2 & \underline{11.3} & \underline{43.1} & 40.1 \\
InternLM2-Chat-7B & \underline{70.7} & \underline{23.6} & 9.5 & 38.0 & \underline{42.3} \\
\multicolumn{6}{c}{\bluecolor{\textit{$\heartsuit$ $\sim20B Models$}}} \\
Llama2-13B-Chat & 43.1 & 5.2 & 1.1 & 6.7 & 15.7 \\
Baichuan2-13B-Chat & 56.0 & 4.3 & 3.4 & 28.3 & 29.1 \\
Mixtral-8x7B-Instruct-v0.1 & 71.7 & 22.5 & 9.6 & 27.2 & 44.3 \\
Qwen-14B-Chat & 57.7 & 27.6 & 8.1 & 47.1 & 40.6 \\
% Qwen1.5-14B-Chat & 63.5 & 24.0 & 10.5 & \textbf{52.0} & \textbf{50.8} \\
\midrule
InternLM2-Chat-20B-SFT & 77.1 & 31.9 & 13.6 & 47.5 & \textbf{50.6} \\
InternLM2-Chat-20B & \textbf{79.6} & \textbf{32.4} & \textbf{14.1} & \textbf{48.0} & 48.5 \\
\hline
\end{tabular}}
\caption{\textbf{Comparison of Chat Models on Math Tasks}. Models are categorized by their parameter size and type, highlighting top performers in each category with \textbf{bold} for overall leaders and \underline{underline} for leaders within their parameter group.}
\vspace{-2mm}
\label{tab:chat_math_result}
\end{table*}

%% file: sections/evaluation/long_context_subsub.tex
\paragraph{Long-context Understanding and Reasoning. }
We mainly evaluate the long-context modeling capability of InternLM2 on the following two benchmarks: L-Eval~\citep{DBLP:journals/corr/abs-2307-11088} and LongBench~\citep{DBLP:journals/corr/abs-2308-14508}. 

\noindent \textbf{L-Eval. } 
L-Eval is a long-context benchmark consisting of 18 subtasks\footnote{
We adopt the first version of L-Eval, corresponding to \href{https://arxiv.org/abs/2307.11088v1}{https://arxiv.org/abs/2307.11088v1}}, 
including texts from various fields such as law, economy, and technology.
L-Eval consists of 411 documents and over 2000 test cases, with an average document length of 7217 words. 
Subtasks in this dataset can be categorized into two major classes:
5 close-ended tasks and 13 open-ended categories. 
Closed-ended tasks are evaluated using exact matching based accuracies,
while open-ended tasks adopt the Rouge score as the metric. 

\noindent \textbf{LongBench. }
LongBench is a long-context benchmark consisting of 21 subtasks with a total of 4750 test cases. 
It is the first bilingual long-context benchmark, with an average English text length of 6711 words and an average Chinese text length of 13386 characters. 
The 21 subtasks are divided into 6 types, providing a more comprehensive evaluation of the model's capabilities in various aspects.

\noindent \textbf{Evaluation Results. }
We report the evaluation results of InternLM2 on long-context benchmarks in \Cref{tab:longtext_chat_results}.
All variants of InternLM2 have demonstrated strong long-context modeling performance across two benchmarks.
InternLM2-Chat-20B-SFT achieves the best performance on L-Eval and outperforms its counterparts by a considerable margin. 
On LongBench, InternLM2-Chat-7B-SFT outperforms other $\leq 7$B Models models across 4 out of 6 subtask categories. 
It obtains a 48.1 overall score, which is only slightly inferior to the 48.4 overall score of ChatGLM3-6B. 
Meanwhile, we noticed that for InternLM2, different parameter sizes do not lead to significant difference in long-context performance, which will be further investigated.

\begin{table*}[!thb]
\centering

\resizebox{\linewidth}{!}{
\tablestyle{10pt}{1.2}
\begin{tabular}{l|ccc|ccccccc}
\toprule
& \multicolumn{3}{c|}{L-Eval} & \multicolumn{7}{c}{LongBench} \\
\hline
Model & Close & Open  & Avg.  & Single & Multi & Summ & FSL & Syn & Code & Avg. \\ \hline

\multicolumn{11}{c}{\greencolor{\textit{ $\leq 7$B Models}}} \\ 
Llama2-7B-Chat & 20.0 & 33.1 & 29.4 & 23.1 & 17.0 & 18.5 & 7.0 & 5.0 & 15.4 & 14.3 \\
Mistral-7B-Instruct-v0.2 & 54.3 & 34.0 & 39.7 & 31.3 & 26.4 & 21.8 & 46.6 & 59.2 & 44.8 & 38.3 \\
Baichuan2-7B-Chat & 36.1 & 32.4 & 33.4 & 30.3 & 18.4 & 21.8 & 42.5 & 6.6 & 36.0 & 25.9 \\
Qwen-7B-Chat & 13.6 & 32.6 & 27.3 & 27.9 & 14.2 & 21.0 & 21.8 & 8.3 & 28.9 & 20.4 \\
% Qwen1.5-7B-Chat & 54.2 & 38.1 & 42.6 & 42.7 & 36.1 & 16.8 & 36.7 & 30.7 & 46.7 & 35.0 \\
ChatGLM3-6B & 59.1 & 35.0 & 41.7 & 40.9 & 45.0 & 26.0 & 56.5 & 65.0 & \underline{57.1} & \textbf{48.4} \\
InternLM2-Chat-7B & 68.6 & \underline{40.8} & 48.5 & 45.7 & 43.1 & \textbf{26.5} & 58.3 & 66.3 & 36.4 & 46.1 \\
InternLM2-Chat-7B-SFT & \underline{68.7} & \underline{40.8} & \underline{48.6} & \textbf{47.3} & \underline{45.2} & 25.3 & \textbf{59.9} & \underline{67.2} & 43.5 & 48.1 \\

\multicolumn{11}{c}{\bluecolor{\textit{ $13 \sim 20$B Models}}} \\ 
Llama2-13B-Chat & 27.3 & 34.4 & 32.5 & 18.2 & 9.9 & 18.5 & 6.7 & 10.0 & 18.1 & 13.6 \\
Mixtral-8x7B-Instruct-v0.1 & 65.3 & 35.6 & 43.9 & 35.0 & 25.6 & 17.1 & 35.9 & 68.2 & 40.0 & 37.0 \\
Baichuan2-13B-Chat & 40.6 & 32.3 & 34.6 & 34.3 & 29.1 & 21.3 & 45.5 & 4.0 & 51.4 & 30.9 \\
Qwen-14B-Chat & 35.3 & 33.5 & 34.0 & 32.6 & 19.4 & 22.5 & 36.9 & 23.7 & 42.9 & 29.7 \\
% Qwen1.5-14B-Chat & 64.2 & 36.4 & 44.1 & 41.7 & 38.9 & 17.2 & 36.9 & 62.3 & \textbf{60.3} & 42.9 \\
InternLM2-Chat-20B & 68.6 & 40.6 & 48.4 & \underline{46.9} & 46.7 & \underline{26.0} & 49.1 & 67.3 & 32.6 & 44.8 \\
InternLM2-Chat-20B-SFT & \textbf{68.8} & \textbf{42.0} & \textbf{49.4} & 46.8 & \textbf{48.7} & 25.6 & \underline{51.2} & \textbf{67.9} & 40.4 & \underline{46.8} \\ \bottomrule
\end{tabular}}
\caption{\textbf{Comparison of Chat Models on Long-Context Benchmarks}. \textbf{Bold} indicates the top performer; 
an \underline{underline} signifies the leading model within the same parameter size group. 
For each benchmark and task group, we report the average accuracies of intra-group subtasks.
\textbf{L-Eval abbrs}: Close $\rightarrow$ close-ended; Open $\rightarrow$ open-ended. 
\textbf{LongBench abbrs}: Single $\rightarrow$ single-document; Multi $\rightarrow$ multi-document; Summ $\rightarrow$ summarization; FSL $\rightarrow$ few shot learning; Syn $\rightarrow$ synthetic. 
All results are obtained with the OpenCompass~\citep{2023opencompass} evaluation toolkit. }
% \vspace{-2mm}
\label{tab:longtext_chat_results}
\end{table*}

\begin{figure*}[!t]
    \centering
    \includegraphics[scale=0.5]{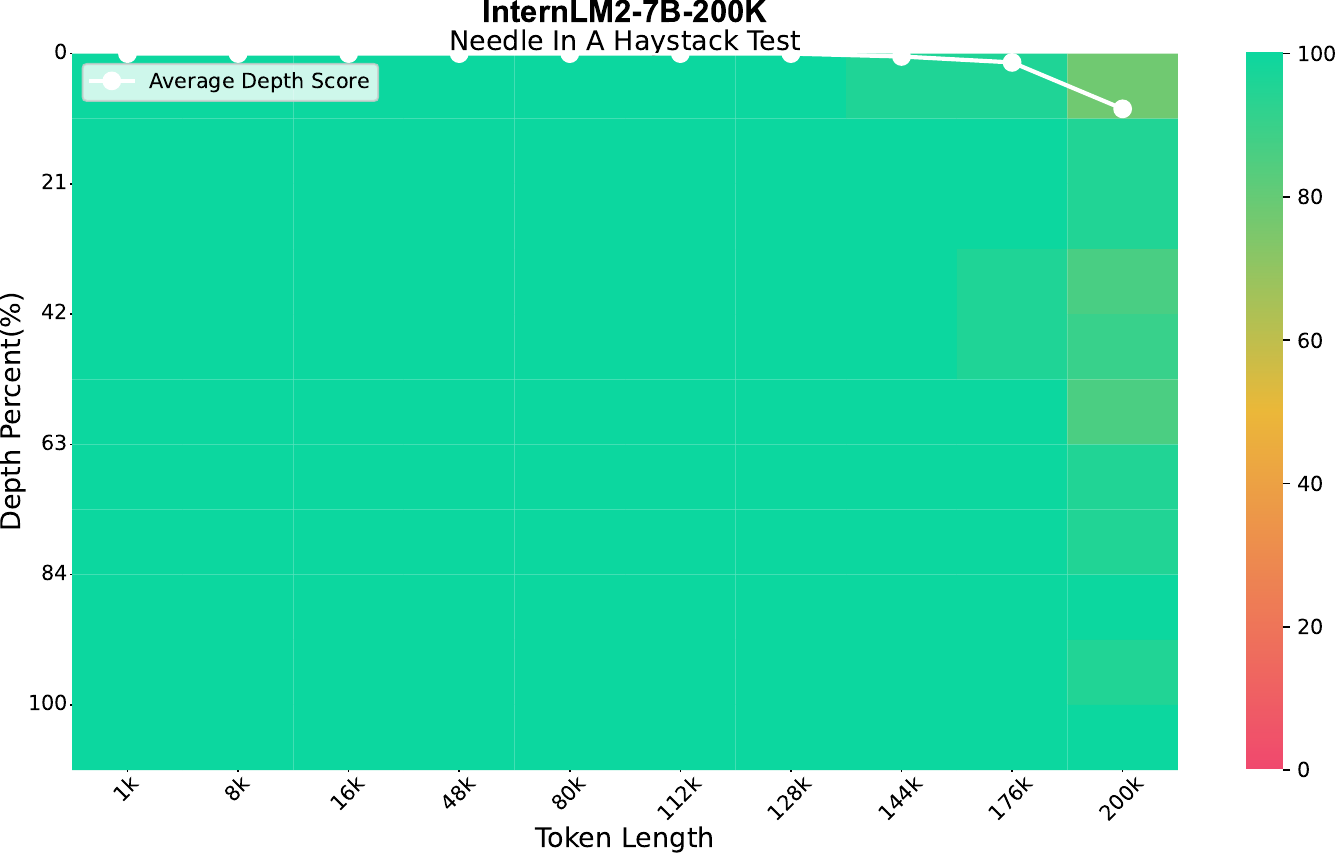}
    \caption{Results on Needle in the Haystack(Chinese).}
    \label{fig:needle_bench}
\end{figure*}

\paragraph{Needle-in-the-Haystack}

``Needle-in-the-Haystack" is a single-needle retrieval task, which is designed to test the Large Language Models' (LLMs) ability to recall a single piece of key information. This is done by inserting a crucial piece of information into a Haystack text of a target length\footnote{Text token length calculations use the GPT-4 tokenizer} at various positions and then querying the model about this key information at the end of the prompt. This method precisely visualizes LLMs' recall capabilities at different positions within long texts of varying lengths. We design a Chinese Haystack following the original idea, and utilize the Skywork/ChineseDomainModelingEval dataset released by \cite{wei2023skywork}, ensuring diversity and quality in the sources of Chinese texts. This dataset covers a wide range of fields from finance to technology, offering high-quality, up-to-date Chinese articles and providing a stable benchmark for assessing different models' abilities to handle domain-specific long texts. For this experiment, we leverage the LMDeploy\footnote{\href{https://github.com/InternLM/lmdeploy}{https://github.com/InternLM/lmdeploy}} \cite{2023lmdeploy} inference engine to accelerate the inference process. The results presented in Figure \ref{fig:needle_bench} effectively demonstrate InternLM2's capability for long-context modeling.

%% file: sections/evaluation/agent.tex
\subsubsection{Tool Utilization}
It is widely acknowledged that external tools and APIs can significantly enhance the capabilities of LLMs to tackle complex, real-world problems~\citep{DBLP:journals/corr/abs-2304-08354,DBLP:journals/corr/abs-2307-16789,DBLP:journals/corr/abs-2302-04761}. To analyze InternLM2's proficiency in tool utilization, 
we conduct experiments across several benchmark datasets: GSM8K~\citep{DBLP:journals/corr/abs-2110-14168}, Math~\citep{DBLP:conf/nips/HendrycksBKABTS21}, the recently introduced MathBench~\citep{Anonymousmathbench}, T-Eval~\citep{chen2023t}, and the template subset of CIBench~\citep{Anonymouscibench}, all employing the ReAct protocol~\citep{DBLP:conf/iclr/YaoZYDSN023} where LLMs alternate between generating thought processes and executing actions. 
Notably, MathBench consists of 3709 questions that span mathematical concepts from primary to high school levels. This dataset enables a comprehensive evaluation of an LLM's ability to solve math problems. T-Eval~\citep{chen2023t} features human-verified, high-quality question instructions with corresponding step-by-step solutions. It measures an LLM's proficiency with everyday tools such as Google Search and Gaode Map across six different dimensions. CIBench, developed by our team, simulates genuine data analysis scenarios using interactive Jupyter notebooks. It encompasses multiple, consecutive tasks and covers the most commonly used Python modules in data analysis, including Pandas, Numpy, and Pytorch. This custom-made benchmark allows for a thorough assessment of an LLM's comprehensive ability in data analysis. 

\begin{figure*}[!t]
    \centering
    \includegraphics[scale=0.38]{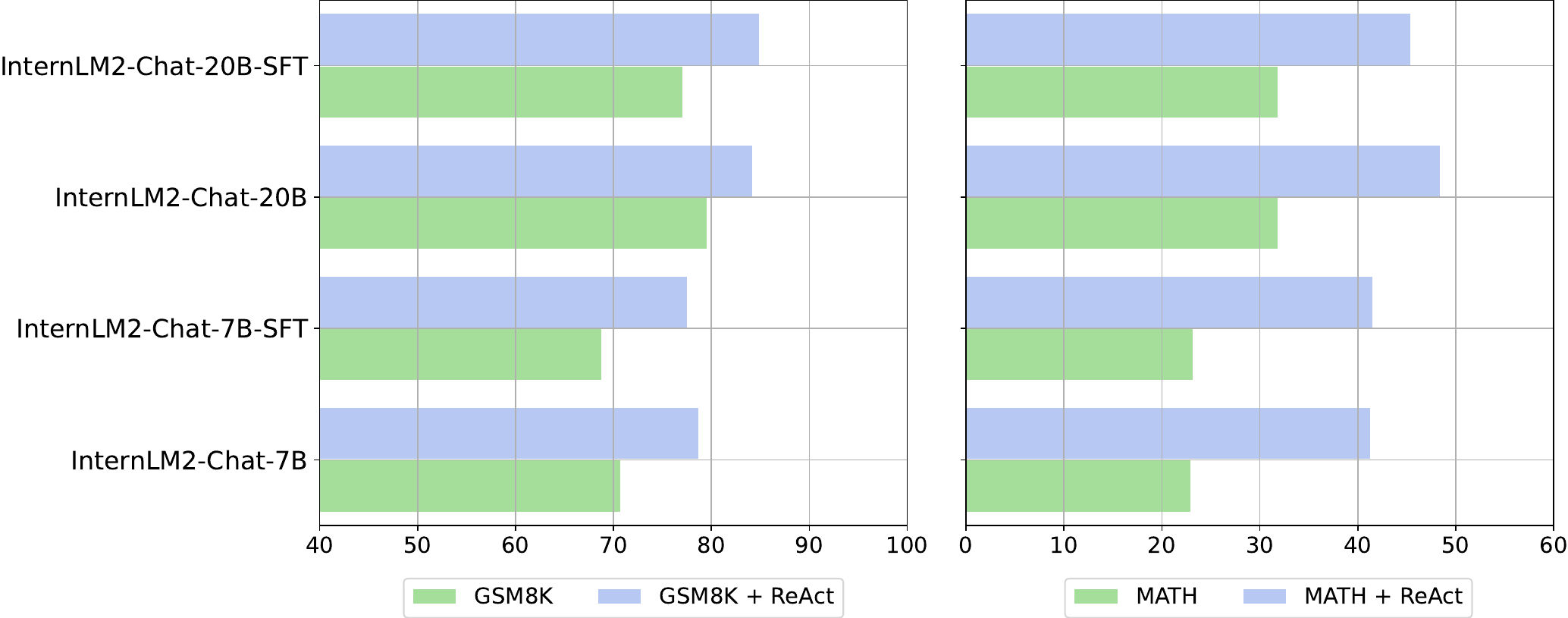}
    \caption{Results on GSM8K (4-shot) and MATH (4-shot) with and w/o Code Interpreter.}
    \label{fig:agent_gsm8k_math}
\end{figure*}

% Notably, MathBench comprises 3709 questions, covering the mathematical concepts in primary, middle, and high school, allowing us to measure LLM's math problem-solving ability thoroughly. T-Eval~\citep{chen2023t} consists of human-verified high-quality question instruction and corresponding step-by-step solutions, measuring LLM's ability on daily used tools, like Google search, Gaode map, etc, from six dimensions. CIBench is constructed by ourselves to simulate authentic scenarios in data analysis, consists of interactive Jupyter notebooks, which contain multiple and consecutive tasks, and cover the most widely-used Python modules used in data analysis (e.g. Pandas, Numpy, Pytorch), benchmark LLM's ability in data analysis comprehensively. 
%The dataset details of CIBench are in the Appendix.
\paragraph{GSM8K, MATH, and MathBench} We utilize the external code interpreter and follow the ReAct protocol to evaluate LLM's ability to solve coding and mathematic problems. The results, as depicted in Figure~\ref{fig:agent_gsm8k_math}, demonstrate a substantial improvement even when using the code interpreter, particularly on the MATH dataset where the enhancement is notably significant. 

\begin{figure*}[!t]
    \centering
    \includegraphics[scale=0.4]{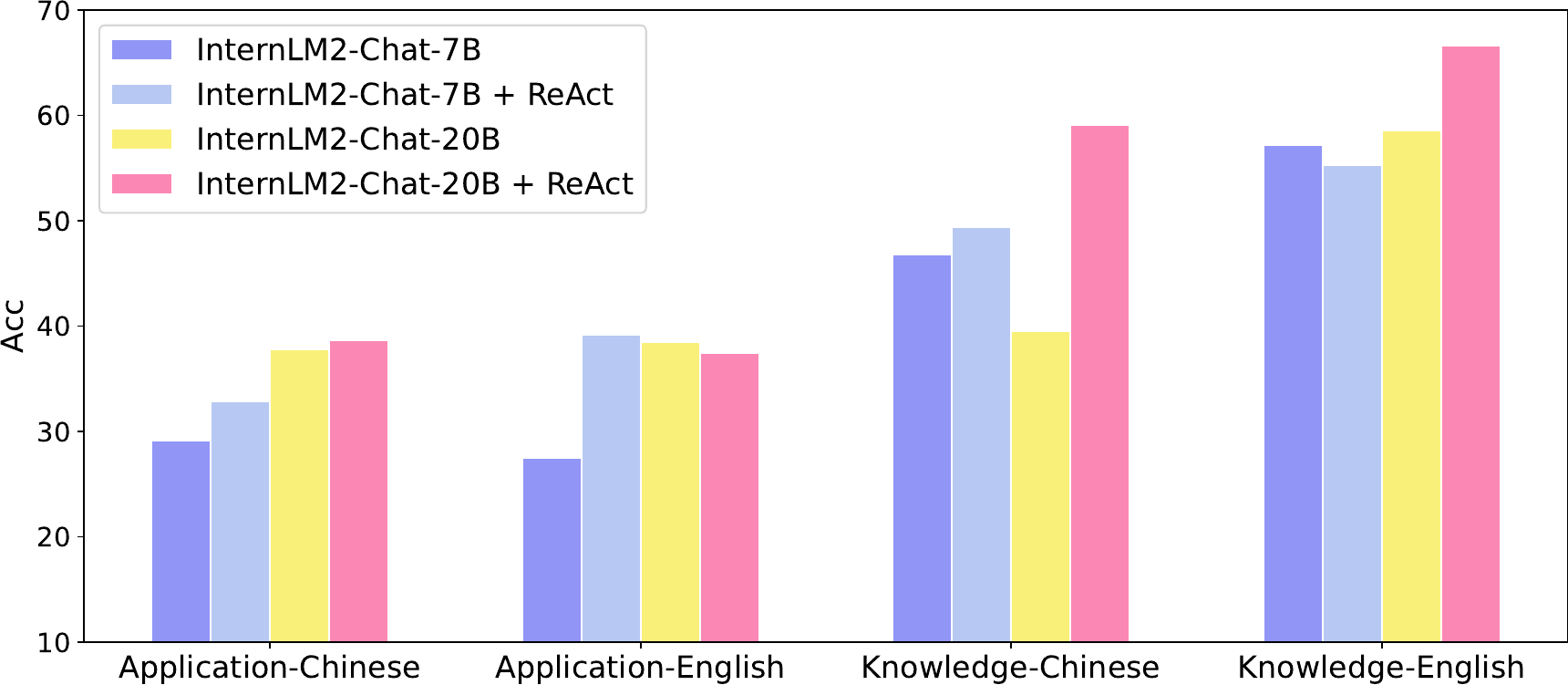}
    \caption{Results on MathBench with and without ReAct.}
    \label{fig:agent_mathbench}
\end{figure*}

% TODO: wait for results to update the text.
As depicted in Figure~\ref{fig:agent_mathbench}, regarding the recently introduced MathBench dataset, the utilization of a code interpreter results in improved performance for InternLM2's performance in most cases, and a minor decrease may be attributed to the incorrect usage of such interpreters. Moreover, notable improvements are observed within the Knowledge domain for InternLM2-20B-Chat and in the Application section for InternLM2-7B-Chat. These disparities can stem from a multitude of factors, including differences in the constitution of their respective training datasets.

% what is the difference of application and knowledge

% TODO: wait for decision on Qwen1.5
% cibench fig:prompt_of_cibench
\paragraph{T-Eval and CIBench} As depicted in Figure~\ref{tab:cibench_teval}, the InternLM2 models consistently demonstrate superior or comparable performance compared to existing models across various benchmarks. Specifically, when comparing models of identical scale, InternLM2-Chat-7B emerges as the top performer on T-Eval and CIBench. Meanwhile, InternLM2-Chat-20B achieves competitive results on T-Eval while obtaining the highest scores on CIBench. Additionally, the InternLM2 series models achieve impressive results in Chinese, showcasing their proficiency in multiple languages.

% Furthermore, InternLM2-Chat observes similar results with InternLM2-Chat-SFT, which serves as compelling evidence for the robustness and effectiveness of our RLHF alignment strategy. This consistency indicates that the fine-tuning process through RLHF has successfully maintained the model's competitive edge without compromising its inherent capabilities.

% Comparing the performance of English with Chinese shows that the performance of Chinese is inferior to English, 

% \paragraph{General Tool Utilization}
% \paragraph{Code Interpreter}

\begin{table*}[!t]

    \setlength{\tabcolsep}{6pt}
    \renewcommand{\arraystretch}{1.1}
    \centering
    \resizebox{0.8\textwidth}{!}{%
    \begin{tabular}{l|ccc|ccc}
    \toprule
    \multirow{2}{*}{\textbf{Model}} & \multicolumn{3}{c|}{\textbf{T-Eval}} & \multicolumn{3}{c}{\textbf{CIBench}}\\ 
    % \cline{2-13}
    % & \multicolumn{2}{c}{\textbf{\textsc{Round Overall}}}
    \addlinespace[0.3pt]
    & \small{English} & \small{Chinese} & \small{Avg} &  \small{English} & \small{Chinese} & \small{Avg}\\ \midrule
    \multicolumn{7}{c}{\greencolor{\textit{ $\leq 7$B Models}}} \\ 
Llama2-7B-Chat   &   37.5 & 37.0 & 37.2 &   12.2   &   10.6   &   11.4   \\ 
% Yi-6B   &   000 & 000 & 000 &   26.8   &   20.9   &   23.8   \\ 
ChatGLM3-6B   &   62.3 & 63.6 & 63.0 &   29.4   &   17.1   &   23.2   \\ 
% DeepSeek-7B   &   000 & 000 & 000 &   27.6   &   22.1   &   24.9   \\ 
% Vicuna-7B   &   000 & 000 & 000 &   28.0   &   18.1   &   23.0   \\
% Baichuan2-7B-Chat &   58.5 & 53.7 &  &   28.0   &   18.1   &   23.0   \\
Qwen-7B-Chat   &   57.5 & 61.1 & 59.3 &   47.8   &   29.2   &   38.5     \\ 
Qwen1.5-7B-Chat   &   54.5 & 50.4 & 52.5 &   44.7   &   26.9   &   35.8   \\ 
Mistral-7B-Instruct-v0.2   &   39.3 & 38.7 & 39.0 &   47.6   &   38.0   &   42.8   \\ 
\midrule
InternLM2-Chat-7B-SFT   &   64.1 & \underline{66.8} & 65.5 &   52.1   &   27.8   &   39.9 \\ 
InternLM2-Chat-7B   &   \underline{\textbf{66.1}} & 65.4 & \underline{65.8}  &   \underline{\textbf{57.1}}   &   \underline{\textbf{40.7}}   &   \underline{\textbf{48.9}}     \\ \midrule
\multicolumn{7}{c}{\bluecolor{\textit{ $13 \sim 20$B Models}}} \\ 
Llama2-13B-Chat   &   48.7 & 44.9 & 46.8 &   20.0   &   12.3   &   16.1     \\ 
% Vicuna-13B   &   000 & 000 & 000 &   42.2   &   37.9   &   40.0   \\ 
Qwen-14B-Chat   &   63.6 & \underline{\textbf{68.7}} & \underline{\textbf{66.2}}  &   54.8   &   40.8   &   47.8   \\ 
Qwen1.5-14B-Chat   &   67.6 & 63.3 & 65.4 &   57.9   &   49.7   &   53.8   \\ 
Mistral-8x7B-Instruct-v0.1   &   58.6 & 58.0 & 58.3 &   42.6   &   40.3   &   41.5   \\ 
\midrule
InternLM2-Chat-20B-SFT   &   64.7 & 64.1 & 64.4 &   49.0   &   43.7   &   46.4   \\ 
InternLM2-Chat-20B   &   \underline{65.6} &  62.9 & 64.3 &   \underline{\textbf{56.0}}   &   \underline{\textbf{54.7}}   &   \underline{\textbf{55.4}}   \\  \bottomrule 
\end{tabular}
}
\vspace{-0.5em}
    \caption{\textbf{Results on T-Eval and CIBench.} \textbf{Bold} indicates the top performer; 
an \underline{underline} signifies the leading model within the same parameter size group. }
    \label{tab:cibench_teval}
\end{table*}

% \paragraph{CIBench}

%% file: sections/evaluation/contamination_table.tex
% 基座模型
\begin{table*}[!thb]
\centering

\resizebox{0.65\linewidth}{!}{
%\tablestyle{10pt}{1.0}
\begin{tabular}{l|ccccc}
\toprule
\textbf{Model} & \(L_{test}\) & \(L_{train}\) & \(L_{ref}\) & \(\Delta_1\) & \(\Delta_2\) \\
\hline
\multicolumn{6}{c}{\greencolor{\textit{  $\leq 7$B Models}}} \\
Llama2-7B & 1.49 & 1.5 & 1.49 & 0 & -0.01 \\
Mistral-7B-v0.1 & 1.43 & 1.44 & 1.41 & 0.02 & -0.01 \\
Baichuan2-7B-Base & 1.47 & 1.48 & 1.45 & 0.02 & -0.01 \\
Qwen-7B & 1.33 & 0.78 & 1.2 & 0.13 & \cellcolor[HTML]{DFDFDF} 0.55 \\
% Qwen1.5-7B & 1.37 & 0.5 & 1.35 & 0.02 & \cellcolor[HTML]{DFDFDF} 0.87 \\
ChatGLM3-6B-Base & 1.3 & 1.16 & 1.35 & -0.05 & 0.14 \\
\midrule
InternLM2-7B-Base & 1.73 & 1.75 & 1.64 & 0.09 & -0.02 \\
InternLM2-7B & 1.48 & 1.14 & 1.46 & 0.02 & \cellcolor[HTML]{DFDFDF} 0.34 \\
\hline
\multicolumn{6}{c}{\bluecolor{\textit{ $13\sim 20$B Models}}} \\
Llama2-13B & 1.42 & 1.42 & 1.45 & -0.03 & 0 \\
Mixtral-8x7B-v0.1 & 1.34 & 1.35 & 1.35 & -0.01 & -0.01 \\
Baichuan2-13B-Base & 1.13 & 0.76 & 1.19 & -0.06 & \cellcolor[HTML]{DFDFDF} 0.37 \\
Qwen-14B & 1.15 & 0.5 & 1.27 & -0.12 &\cellcolor[HTML]{DFDFDF}  0.65 \\
% Qwen1.5-14B & 1.28 & 0.45 & 1.34 & -0.06 & \cellcolor[HTML]{DFDFDF} 0.83 \\
\midrule
InternLM2-20B-Base & 1.66 & 1.67 & 1.58 & 0.08 & -0.01 \\
InternLM2-20B & 1.52 & 1.17 & 1.48 & 0.04 & \cellcolor[HTML]{DFDFDF} 0.35 \\
\bottomrule
\end{tabular}}
\caption{\textbf{Evaluation of Base Models on GSM8K Contamination}.}
% \vspace{-2mm}
\label{tab:language_conta_results}
\end{table*}

%% file: sections/appendix.tex
\section{Appendix}
\subsection{Acknowledgements}
The InternLM project has been made possible thanks to the contributions of the following individuals. We would like to express our gratitude to them for their support:\footnote{Authors are ordered alphabetically by the last name.}:

Xiang Cao, Yuanyuan Cao, Weihan Cao, Yuhang Cao, Xiaodong Chang, Wenjun Chen, Lu Chen, Songyang Gao, Jianfei Gao, Yihan Geng, Honglin Guo, Ziwei Ji, Tian Lan, Menglei Li, Peiji Li, Bowen Li, Mo Li, Xingpu Li, Yiqi Lin, Lindong Lu, Han Lv, Ren Ma, Yichuan Ma, Xiuda Ma, Jiahui Peng, Runyu Peng, Wenwen Qu, Guanlin Shen, Haowen Shen, Jin Shi, Yixin Song, Chenlin Su, Xiaohui Su, Zhewen Tan, Shihan Tian, Zhongying Tu, Bo Wang, Shasha Wang, Zhiqiang Wang, Zerui Wang, Chonghua Wang, Mengke Wang, Jiang Wu, Hui Wu, Shuhao Xing, Rui Xu, Qian Yao, Chang Yuan, Zhiyuan Zeng, Zijian Zhang, Yifan Zhang, Min Zhang, Zhiyuan Zhao, Ying Zhao, Zilong Zheng, Yunhua Zhou, Zhihao Zhu, Jie Zhu, Lin Zhu.

\subsection{Prompts for Evaluation}

\begin{figure*}[h]
    \centering
    \includegraphics[width=1.0\columnwidth]{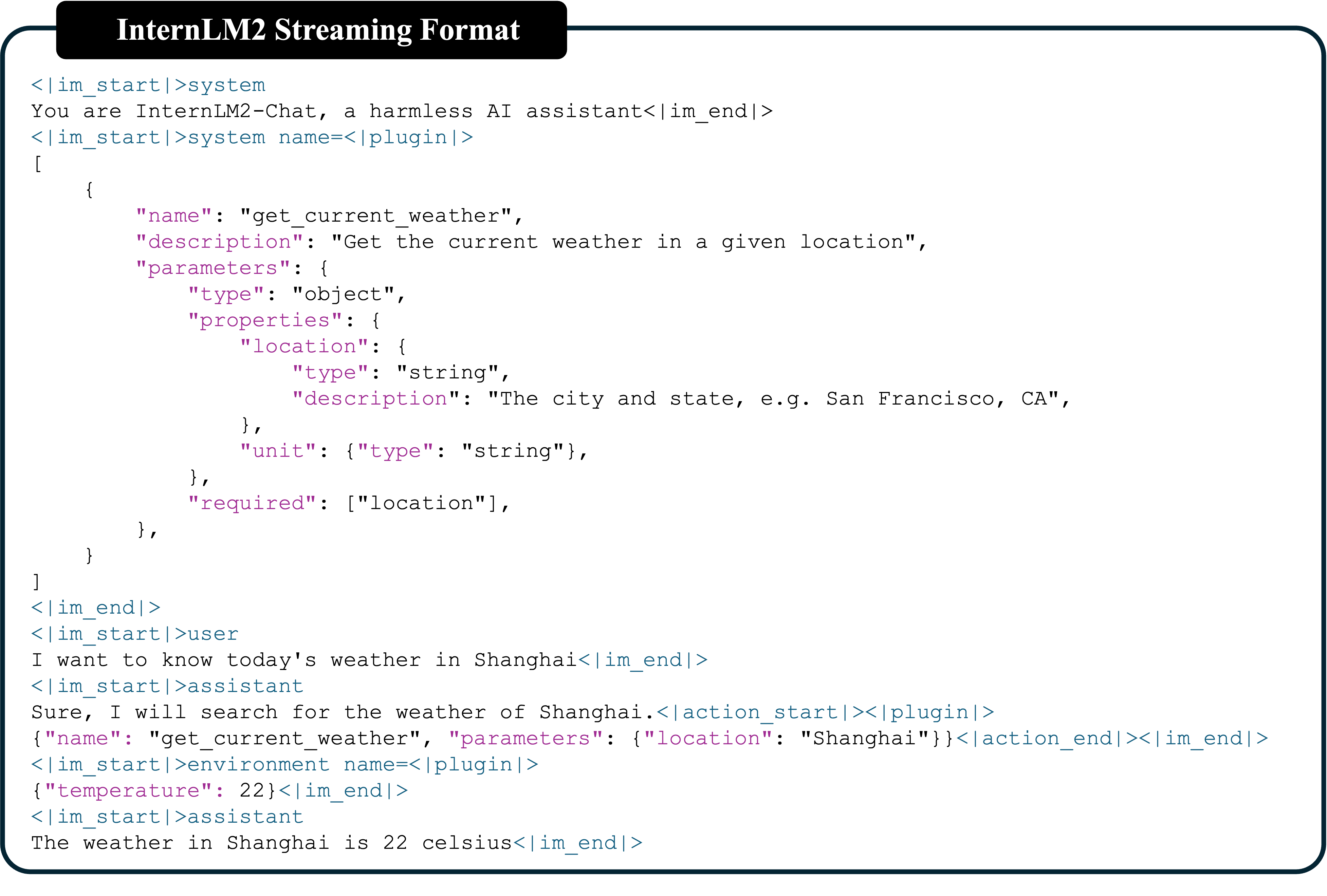}
    \caption{Illustraction of streaming format adopted by InternLM2-Chat to conduct function calls, especially JSON format to ease downstream applications.}
    \label{fig:func_call}
\end{figure*}

\begin{figure*}[!ht] 
\vspace{-5mm}
\begin{AIbox}{Prompts Example}
{\color{blue}\bf Prompt:} \\
{
    \textbf{Question:} 
    Let $R$ be a ring and let $U$ and $V$ be (two-sided) ideals of $R$. Which of the following must also be ideals of $R$?   
   
}

\textbf{Error Analysis:} \\
In crafting its response, the model fails to accurately grasp the concept of an ideal within a ring.

\end{AIbox} 
\caption{An example of misunderstandings of mathematical concepts.}
\label{fig: misunderstandings of mathematical concepts}
\end{figure*}

\input{sections/evaluation/agent_appendix}

\input{sections/evaluation/alignment_appendix}

%% file: sections/evaluation/agent_appendix.tex
\begin{figure*}[!ht] 
\vspace{-5mm}
\begin{AIbox}{Prompt in CIBench}
% {\color{blue}\bf Prompt:} \\
% {
%     \textbf{Question:} 
%     Let $R$ be a ring and let $U$ and $V$ be (two-sided) ideals of $R$. Which of the following must also be ideals of $R$?   
   
% }

% \textbf{Error Analysis:} \\
% In crafting its response, the model fails to accurately grasp the concept of an ideal within a ring.
{\color{blue}\bf System Prompt:}

You are an assistant who can utilize external tools.

\texttt{IPythonInterpreter}: It can run Python code in a manner as jupyter notebook. The code must be a valid code that contains only python method.

% \lstset{style=prompt_json}
% \begin{lstlisting}
% \{\`\ IPythonInterpreter\`\: \`\It can run Python code in a manner as jupyter notebook. The code must be a valid code that contains only python method.\`\ \}
% \end{lstlisting}

To use a tool, please response with the following format: \\

\textbf{Thought}: Think what you need to solve, do you need to use tools? \\

\textbf{Action}: The tool name, should be one of \texttt{IPythonInterpreter}. \\

\textbf{Action Input}: The input to the tool that you want to use. \\
\\
The tool will give you response after your response using the following format: \\

\textbf{Response}: the results after call the tool. \\
\\
Therefore DO NOT generate tool response by yourself. \\

Also please follow the guidelines: \\
1. Always use code interpreter to solve the problem. \\
2. The generated codes should always in a markdown code block format. \\
3. The generated codes will be executed in an ipython manner and the results will be cached. \\
4. Your responded code should always be simple and only solves the problem in current step. \\

For example: \\

File url: \texttt{xxxx} \\
\#\#\# Step 1. Load the dataset from the url into a pandas DataFrame named \texttt{df}. \\

\textbf{Thought}: We should use \texttt{pandas} to solve this step. \\
\textbf{Action}: \texttt{IPythonInterpreter} \\
\textbf{Action Input}: 

\lstset{style=prompt_json}
\begin{lstlisting}
import pandas as pd \\
url = "xxxx" \\
data = pd.read\_csv(url) \\
\end{lstlisting}

\textbf{Response}: The code is succeed without any outputs.

Let us begin from here! \\

{\color{blue}\bf User Prompt:} \\

\{Question\}. Please use \{modules\} modules.

\end{AIbox} 
\caption{Prompt used in CIBench}
\label{fig:prompt_of_cibench}
\end{figure*}

\begin{figure*}[!ht] 
\vspace{-5mm}
\begin{AIbox}{Prompt in CIBench}
{\color{blue}\bf System Prompt:}

You are an assistant who can utilize external tools. \\
\{tool\_description\} \\
To use a tool, please use the following format: \\

\lstset{style=prompt_json}
\begin{lstlisting}
\{thought\}Think what you need to solve, do you need to use tools? \\
\{action\}the tool name, should be one of [\{action\_names\}] \\
\{action\_input\}the input to the action \\
\end{lstlisting}

The response after utilizing tools should using the following format: \\

\lstset{style=prompt_json}
\begin{lstlisting}
\{response\}the results after call the tool. \\
\end{lstlisting}

If you already know the answer, or you do not need to use tools, \\
please using the following format to reply: \\

\lstset{style=prompt_json}
\begin{lstlisting}
\{thought\}the thought process to get the final answer \\
\{finish\}final answer \\
\end{lstlisting}

Begin! \\

{\color{blue}\bf Few-shot Prompt:} \\

\textbf{HUMAN}: Find the coefficient of $x^3$ when $3(x^2 - x^3+x) +3(x +2x^3- 3x^2 + 3x^5+x^3) -5(1+x-4x^3 - x^2)$ is simplifie.

\textbf{BOT}: 

\textbf{Tool}:\texttt{PythonInterpreter}

\textbf{Tool Input}:

\lstset{style=prompt_json}
\begin{lstlisting}
from sympy import symbols, simplify

def solution(): \\
\quad x = symbols('x') \\
\quad expr = $3*(x**2 - x**3 + x) + 3*(x + 2*x**3 - 3*x**2 + 3*x**5 + x**3) - 5*(1 + x - 4*x**3 - x**2)$ \\
\quad simplified\_expr = simplify(expr) \\ 
\quad x3\_coefficient = simplified\_expr.as\_coefficients\_dict()[x**3] \\ 
\quad result = x3\_coefficient \\ 
\quad return result \\
\end{lstlisting}

\textbf{SYSTEM}: Response:26 \\

\textbf{BOT}: FinalAnswer: The final answer is $26$. I hope it is correct. \\
...
\end{AIbox} 
\caption{Prompt used in MATH}
\label{fig:prompt_of_cibench}
\end{figure*}

\begin{figure*}[!ht] 
\vspace{-5mm}
\begin{AIbox}{Prompt in CIBench}
{\color{blue}\bf System Prompt:}

You are an assistant who can utilize external tools. \\
\{tool\_description\} \\
To use a tool, please use the following format: \\

\lstset{style=prompt_json}
\begin{lstlisting}
\{thought\}Think what you need to solve, do you need to use tools? \\
\{action\}the tool name, should be one of [\{action\_names\}] \\
\{action\_input\}the input to the action \\
\end{lstlisting}

The response after utilizing tools should using the following format: \\

\lstset{style=prompt_json}
\begin{lstlisting}
\{response\}the results after call the tool. \\
\end{lstlisting}

If you already know the answer, or you do not need to use tools, \\
please using the following format to reply: \\

\lstset{style=prompt_json}
\begin{lstlisting}
\{thought\}the thought process to get the final answer \\
\{finish\}final answer \\
\end{lstlisting}

Begin! \\

{\color{blue}\bf Few-shot Prompt:} \\

\textbf{HUMAN}: 
Mark's basketball team scores 25 2 pointers, 8 3 pointers and 10 free throws.  Their opponents score double the 2 pointers but half the 3 pointers and free throws.  What's the total number of points scored by both teams added together?

\textbf{BOT}: 

\textbf{Tool}:\texttt{PythonInterpreter}\\

\textbf{Tool Input}: 
\lstset{style=prompt_json}
\begin{lstlisting}
def solution():\\   
mark\_pointers\_2 = 25 * 2   
mark\_pointers\_3 = 8 * 3\\   
mark\_free\_throws = 10 * 1\\   
mark\_points\_scored = mark\_pointers\_2 + mark\_pointers\_3 + mark\_free\_throws\\   
opponents\_pointers\_2 = mark\_pointers\_2 * 2\\   
opponents\_pointers\_3 = mark\_pointers\_3 / 2\\   
opponents\_free\_throws = mark\_free\_throws / 2\\   
opponents\_points\_scored = opponents\_pointers\_2 + opponents\_pointers\_3 + opponents\_free\_throws\\   
total\_points\_scored = mark\_points\_scored + opponents\_points\_scored\\   
result = total\_points\_scored\\   
return result \\
\end{lstlisting}

\textbf{SYSTEM}: Response:201 \\

\textbf{BOT}: Thought: According to the response, I got the answer \\ FinalAnswer: 201 \\

...
\end{AIbox} 
\caption{Prompt used in GSM8K}
\label{fig:prompt_of_agent_gsm8k}
\end{figure*}

% \begin{lstlisting}[style=code]
% # Detect corners using Shi-Tomasi corner detector
% corners = cv2.goodFeaturesToTrack(equalized_image, 
%                                   maxCorners=max_corners, 
%                                   qualityLevel=quality_level, 
%                                   minDistance=min_distance, 
%                                   blockSize=block_size)
% # Mark the corners with circles on the image
% marked_image = equalized_image.copy()
% for corner in corners:
%     x, y = corner.ravel()
%     cv2.circle(marked_image, (int(x), int(y)), 5, (255, 0, 0), -1)  # Draw a blue filled circle at each corner

% # Show the marked image
% plt.imshow(marked_image, cmap='gray')
% plt.axis('off')  # Turn off axis numbers and ticks
% plt.show()

% \end{lstlisting}

%% file: sections/evaluation/alignment_appendix.tex
\begin{figure*}[!ht] \label{appendix_alpacaeval}
\vspace{-5mm}
\begin{AIbox}{Prompt in AlpacaEval}
% {\color{blue}\bf Prompt:} \\
% {
%     \textbf{Question:} 
%     Let $R$ be a ring and let $U$ and $V$ be (two-sided) ideals of $R$. Which of the following must also be ideals of $R$?   
   
% }

% \textbf{Error Analysis:} \\
% In crafting its response, the model fails to accurately grasp the concept of an ideal within a ring.
{\color{blue}\bf System Prompt:}\\
You are a highly efficient assistant, who evaluates and selects the best large language model (LLMs) based on the quality of their responses to a given instruction. This process will be used to create a leaderboard reflecting the most accurate and human-preferred answers. \\

{\color{blue}\bf User Prompt:} \\
I require a leaderboard for various large language models. I'll provide you with prompts given to these models and their corresponding outputs. Your task is to assess these responses, and select the model that produces the best output from a human perspective.\\

\textbf{Instruction}
\lstset{style=prompt_json}
\begin{lstlisting}
{
    "instruction": """{instruction}"""
}
\end{lstlisting}

\textbf{Model Outputs}

Here are the unordered outputs from the models. Each output is associated with a specific model, identified by a unique model identifier.\\
\begin{lstlisting}
{
    {
        "model_identifier": "m",
        "output": """{output_1}"""
    },
    {
        "model_identifier": "M",
        "output": """{output_2}"""
    }
}
\end{lstlisting}
\textbf{Task}

Evaluate the models based on the quality and relevance of their outputs, and select the model that generated the best output. Answer by providing the model identifier of the best model. We will use your output as the name of the best model, so make sure your output only contains one of the following model identifiers and nothing else (no quotes, no spaces, no new lines, ...): m or M.\\

\textbf{Best Model Identifier}

\end{AIbox} 
\caption{Prompt used in AlpacaEval}
\label{fig:prompt_of_alpacaeval}
\end{figure*}

\begin{figure*}[!ht] 
\vspace{-5mm}
\begin{AIbox}{Prompt in MTBench}
{\color{blue}\bf System Prompt:}\\
Please act as an impartial judge and evaluate the quality of the response provided by an AI assistant to the user question displayed below. Your evaluation should consider factors such as the helpfulness, relevance, accuracy, depth, creativity, and level of detail of the response. You evaluation should focus on the assistant's answer to the second user question. Begin your evaluation by providing a short explanation. Be as objective as possible. After providing your explanation, you must rate the response on a scale of 1 to 10 by strictly following this format: \textbf{[[rating]]}, for example: \textbf{Rating: [[5]]} \\
{\color{blue}\bf User Prompt:} \\
\lstset{style=prompt_json}
\begin{lstlisting}
<|The Start of Assistant A's Conversation with User|>

### User:
{question_1}

### Assistant A:
{answer_1}

### User:
{question_2}

### Assistant A:
{answer_2}

<|The End of Assistant A's Conversation with User|>
\end{lstlisting}
\end{AIbox} 
\caption{Prompt used in MTBench}
\label{fig:prompt_of_MTBench}
\end{figure*}